\documentclass[5p,times,twocolumn]{elsarticle}

\usepackage[ruled,vlined]{algorithm2e}
\usepackage[table]{xcolor}
\usepackage[most]{tcolorbox}
\usepackage{subcaption}
\usepackage{xurl}
\usepackage{booktabs}
\usepackage{array}
\usepackage{dblfloatfix}
\usepackage[section]{placeins}
\usepackage{todonotes}\usepackage{setspace}
\usepackage{amsmath,amsfonts,amssymb}
\usepackage{multicol}
\usepackage{balance}
\balance

\usepackage{pifont}
\newcommand{\cmark}{\ding{51}}
\usepackage{siunitx}
\def\tsc#1{\csdef{#1}{\textsc{\lowercase{#1}}\xspace}}
\tsc{WGM}
\tsc{QE}

\newcommand{\alma}{\texttt{Alma}}

\newcommand{\fmn}{\texttt{FMN}}
\newcommand{\ddn}{\texttt{DDN}}
\newcommand{\pdpgd}{\texttt{PDPGD}}
\newcommand{\pdgd}{\texttt{PDGD}}
\newcommand{\sigmazero}{\texttt{$\sigma$-zero}}
\newcommand{\APGDmin}{\texttt{APGD}$_{min}$}

\newcommand{\CE}{\texttt{CE}}

\newcommand{\DLR}{\texttt{DLR}}

\newcommand{\vct}[1]{\ensuremath{\boldsymbol{#1}}}
\newcommand{\A}{\mathcal{A}}                       
\newcommand{\ens}{\mathcal{E}}                     
\newcommand{\ensopt}{\widehat{\ens}}                
\newcommand{\DD}{\mathcal{D}}
\newcommand{\MM}{\mathcal{M}}
\DeclareMathOperator*{\argmax}{arg\,max}

\newcommand{\attackoptimality}{Attack Optimality Index\xspace}
\newcommand{\gaoi}{\ensuremath{\overline{\mathrm{AOI}}}\xspace}
\newcommand{\defenseoptimality}{Defense Optimality Index\xspace}
\newcommand{\aoi}{\text{AOI}\xspace}
\newcommand{\doi}{\text{DOI}\xspace}

\newcommand{\myparagraph}[1]{\noindent\textbf{#1}}
\newcommand{\ie}{i.e.\xspace}

\SetKwInput{KwInput}{Input}
\SetKwInput{KwOutput}{Output}
\SetKwProg{Fn}{procedure}{}{}
\SetKwComment{Comment}{$\triangleright$\ }{}

\SetCommentSty{algcommentfont}
\SetKwComment{tcp}{\(\triangleright\)\ }{}

\definecolor{pastelgreen}{RGB}{205,240,205}
\definecolor{pastelorange}{RGB}{255,230,200}
\definecolor{pastelblue}{RGB}{200,220,250}
\definecolor{pastelpurple}{RGB}{230,210,255}

\definecolor{bestbg}{HTML}{D9EAD3}
\definecolor{worstbg}{HTML}{F4CCCC}
\newcommand{\hlbest}[1]{\textcolor{green!60!black}{#1}}

\definecolor{ganavy}{HTML}{1F2A44}
\definecolor{gaborder}{HTML}{D9DCE1}
\definecolor{gapanel}{HTML}{F5F6F8}

\providecolor{ganavy}{HTML}{1F2A44}
\providecolor{teal}{HTML}{2A9D8F}
\providecolor{coral}{HTML}{E76F51}
\providecolor{bestbg}{HTML}{D9EAD3}
\providecolor{worstbg}{HTML}{F4CCCC}
\providecommand{\hlbest}[1]{\textcolor{green!60!black}{#1}}

\colorlet{barU}{ganavy!50}   
\colorlet{barR}{teal!60}     
\colorlet{barT}{coral!60}    

\begin{document}

\let\WriteBookmarks\relax
\setcounter{topnumber}{5}
\setcounter{dbltopnumber}{5}
\setcounter{totalnumber}{10}
\renewcommand{\topfraction}{0.95}
\renewcommand{\dbltopfraction}{0.95}
\renewcommand{\textfraction}{0.05}
\renewcommand{\floatpagefraction}{0.85}
\renewcommand{\dblfloatpagefraction}{0.85}



\title{Adversarial Frontiers: Minimum-Norm Attack Ensembles for Robustness Evaluation}  

\author[1,2]{Luca Scionis\fnref{fn1}}

\author[1,2]{Luca Melis\fnref{fn1}}

\author[2]{Maura Pintor\corref{cor1}}
\ead{maura.pintor@unica.it}

\author[2]{Fabio Brau}

\author[2]{Ambra Demontis}

\author[2]{\\Giorgio Fumera}

\author[2,3]{Fabio Roli}

\author[2]{Battista Biggio}

\cortext[cor1]{Corresponding author}
\fntext[fn1]{These authors contributed equally to this work.}

\affiliation[1]{organization={Sapienza University of Rome},
            country={Italy}}

\affiliation[2]{organization={University of Cagliari},
            country={Italy}}

\affiliation[3]{organization={University of Genoa},
            country={Italy}}
            
\begin{abstract}
Adversarial robustness is commonly evaluated with predefined attack ensembles, such as AutoAttack, at a single perturbation budget $\varepsilon$ and on a selective choice of perturbation norms. We argue this formulation is fundamentally limited. First, robustness--perturbation curves may intersect or decay at different rates across models, making single-$\varepsilon$ rankings unstable. Second, current ensembles provide no evidence of optimality, leaving an unknown gap to worst-case performance. Third, fixed attack configurations provide no systematic control over the trade-off between attack strength and evaluation cost. 
To address these limitations, we introduce a unified evaluation framework based on a comprehensive pool of minimum-norm attacks and robustness–perturbation curves across $\ell_0$, $\ell_1$, $\ell_2$ and $\ell_\infty$ norms. We define the attack frontier as the worst-case robustness estimate the attack pool produces against a model. We then formalize evaluation as a frontier-approximation problem, constructing minimum-norm attack ensembles, optimized subsets of the comprehensive pool, that approach the frontier under a controllable query budget, with larger budgets monotonically tightening the estimate. Furthermore, we define the defense frontier as the maximum robustness across the model set at each perturbation size. We finally propose the Defense Optimality Index to rank defenses by their gap to the defense frontier, providing a ranking without selecting a reference $\varepsilon$.
On CIFAR-10 and ImageNet, our ensembles match or exceed AutoAttack on most defenses at every budget tier, at fixed and controllable query cost, offering practitioners a query-controlled, curve-based alternative to fixed-$\varepsilon$ evaluation.
\end{abstract}

\maketitle

\section{Introduction}\label{sec:intro}

\begin{figure}[!t]
    \centering
    \includegraphics[width=0.8\linewidth]{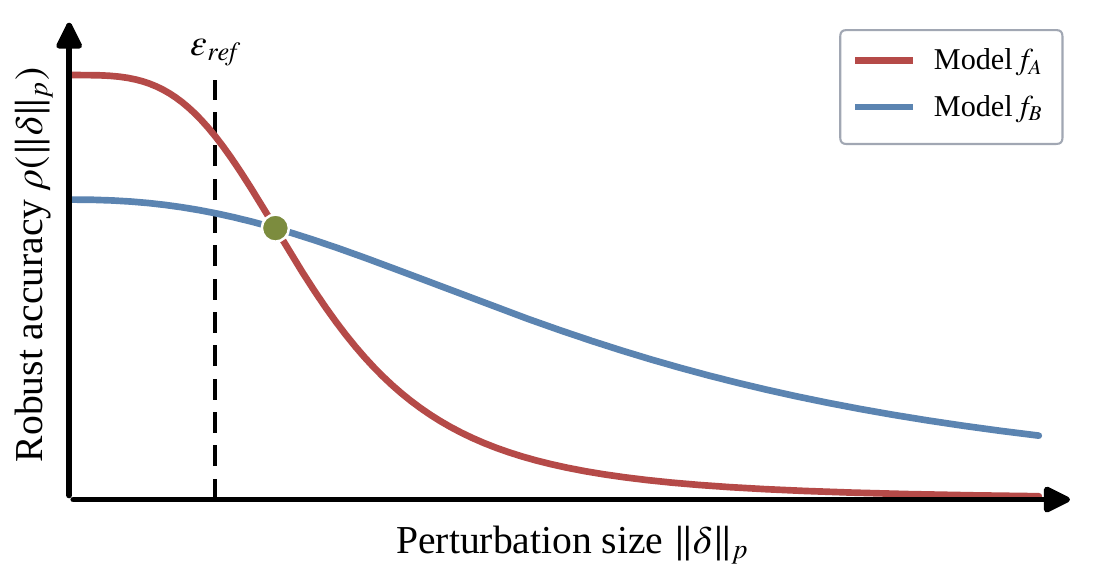}
    \caption{%
        \textbf{The limitation of fixed-$\varepsilon$ evaluations.} 
        This plot shows the robustness--perturbation curves, $\rho(\varepsilon)$, for two hypothetical models, whose robustness decays at different rates.
        At a fixed reference budget $\varepsilon_{\text{ref}}$, the red model is more robust than the blue one. Yet, the latter appear stronger at larger $\varepsilon$. 
        In Section~\ref{sec:experiments} we demonstrate empirically how single-point evaluations are incomplete, motivating the need for $\varepsilon$-free evaluation frameworks.
    }
    \label{fig:fixed_budget_limitation}
\end{figure}

Deep neural networks (DNNs) are vulnerable to adversarial examples~\cite{biggio13-ecml,szegedy2014intriguing}, carefully-perturbed inputs that induce misclassification. 
Reliably measuring how robust a model is to such perturbations is critical for deployment.
The main metric adopted for this evaluation is the \emph{robust accuracy}, \ie, the fraction of inputs correctly classified by a model against adversarial perturbations bounded by a specific $\ell_p$-norm and a maximum perturbation size $\varepsilon$. 
Common evaluation frameworks rely on adversarial attacks to estimate this metric by perturbing each input to search for a misclassifying perturbation. 
Yet, a single weak attack may fail to identify a perturbation even when one exists, overestimating robustness. 
Therefore, state-of-the-art frameworks use ensembles of diverse attacks to compensate for individual failures, yielding more reliable robustness estimates.
The most relevant example adopting this approach is RobustBench~\cite{croce2021robustbench}, a common state-of-the-art benchmark of robust models, whose ranking relies on AutoAttack (AA)~\cite{croce2020reliable}, a fixed ensemble of four sequential attacks. 
This methodology, however, is characterized by the following limitations:

\noindent (i) \textit{Robustness is a function, not a point.}
The complete robustness profile of a model is defined by the \emph{robustness--perturbation curve}, which plots the robust accuracy as a function of the perturbation size $\varepsilon$, for a given $\ell_p$-norm.
Best-practice guidelines have long recommended reporting the complete curve~\cite{biggio2018wild, carlini2019evaluating,risse2021compare}.
However, current benchmarks evaluate models at a single, arbitrary $\varepsilon$.
As shown in Figure~\ref{fig:fixed_budget_limitation}, this is inherently problematic, since robustness--perturbation curves can intersect or decay at different rates.
The selective choice of the perturbation norm further aggravates this problem. 
Each $\ell_p$ norm captures different types of attack.
A model robust to dense perturbations spread across the input ($\ell_2$, $\ell_\infty$ norm) might be substantially weaker to sparse perturbations ($\ell_0$, $\ell_1$). 
Robustness to sparse perturbations is not covered by available benchmarks.

\noindent (ii) \textit{It is not possible to evaluate how close the attack is to the worst case.}
AA is provided with no tools to check how close the robust accuracy it finds is to the empirical optimum (obtained considering also other attacks that are not in AA). 
A high robust accuracy may therefore reflect a strong defense or simply a weak attack, and the two are indistinguishable within the current framework.

\noindent (iii) \textit{The pipeline cannot be adapted by changing the attacks nor their relative cost budgets.} AA is parameter-free and pre-configured by design, but this means evaluation quality and computational cost are coupled and non-adjustable. A practitioner with a small budget cannot obtain an approximate evaluation, while one with a large budget cannot improve beyond the fixed pipeline.

We address the aforementioned limitations with the following contributions: 

\noindent (i) \textit{We introduce a unified evaluation framework based on minimum-norm adversarial attacks and robustness--perturbation curves to characterize robustness across the $\ell_0$, $\ell_1$, $\ell_2$, and $\ell_\infty$ perturbation norms}. 
We call \emph{attack frontier} the worst-case robustness estimate any given combination of attacks can produce. 

\noindent (ii) \textit{We formalize robustness evaluation as a frontier-approximation problem.}
Starting from a pool of minimum-norm attacks and building on the \attackoptimality{} of AttackBench~\cite{cina2025attackbench}, we construct near-optimal attack ensembles, \ie, specific subsets of Attacks taken from the initial pool.
These ensembles are built to approximate the empirical attack frontier as closely as possible under any given query budget, consistently across all the models we evaluated.
Importantly, this approach measures how tight the approximation of the attack frontier is, for each target model.
We demonstrate that increasing the query budget systematically reduces this gap, thereby establishing a principled trade-off between computational cost and the tightness of the approximation.

\noindent (iii) \textit{We introduce the defense frontier and the \defenseoptimality{} (\doi{}) as a defense-side counterpart to the attack frontier.} 
Analogous to the attack frontier, we define the \emph{defense frontier} as the empirical upper bound of robustness achievable across a given pool of evaluated models.
This provides an objective baseline against which any individual defense can be measured.
We then formalize the gap between a model's robustness--perturbation curve and the defense frontier as the \defenseoptimality{} (\doi{}). 
Unlike the robust accuracy at a single $\varepsilon$, the \doi{} aggregates this gap across the entire perturbation range.
This yields a stable, $\varepsilon$-independent ranking of defenses.

\noindent (iv) \textit{We leverage our framework to establish a new, budget-aware evaluation benchmark.} 
Building on our frontier-approximation strategy, we construct attack ensembles approximating the frontier for the $\ell_0$, $\ell_1$, $\ell_2$, and $\ell_\infty$ perturbation norms, at three nested query budget tiers. 
We deploy these ensembles to compute the \doi{} of existing defenses, proposing the resulting 
benchmark as a principled, budget-aware alternative to fixed-pipeline evaluations. 
Evaluation quality is thus decoupled from computational cost, allowing practitioners to trade-off their query budget against the tightness of the robustness estimate.

We evaluate our approach on $13$ held-out defenses on CIFAR-10 and ImageNet. Our minimum-norm attack ensembles match or exceed AA's optimality on the majority of models at every budget tier, with the margin generally improving as queries increase.
The advantage is largest in $\ell_2$, where our ensemble already matches or exceeds AA on $12/13$ defenses at $Q=4\text{k}$ queries, despite AA spending up to $7\,566$ queries at its reference $\varepsilon$. 
In $\ell_1$ it climbs from $8/13$ at $Q=4\text{k}$ to $13/13$ at $Q=12\text{k}$, and in $\ell_\infty$ it reaches $10/13$.
We also show that AA rankings are highly unstable: varying $\varepsilon$ beyond the RobustBench reference reorders nearly every defense. On CIFAR-10, $17$ defenses shift by at least $5$ positions under $\ell_2$, and on ImageNet all $10$ models shift by at least $2$ positions under $\ell_\infty$.
The \doi, instead, produces a single, $\varepsilon$-independent ranking.

The remainder of the paper is organized as follows. 
After reviewing minimum-norm attacks and the attack frontier in Section~\ref{sec:background}, we formalize the budget allocation problem and propose a greedy approach for constructing optimal attack ensembles in Section~\ref{sec:methodology}.
Section~\ref{sec:doi} extends this framework by introducing the defense frontier and the \defenseoptimality{}.
We validate our approach in Section~\ref{sec:experiments}, discuss related work in Section~\ref{sec:related}, and conclude in Section~\ref{sec:conclusion}.

\section{Background}
\label{sec:background}

We first discuss the main tool that should drive~\cite{biggio2018wild,carlini2019evaluating,gopfert2019adversarial,risse2021compare} robustness evaluations: the robustness--perturbation curve. Reliably evaluating a model's robustness requires characterizing how it degrades at increasing perturbation strength. This is what the robustness--perturbation curve captures, reporting the fraction of inputs that remain correctly classified as the allowed perturbation grows.

We then introduce the tools we use to compute the robustness--perturbation curve, \ie, minimum-norm adversarial attacks. Since these attacks are empirical, they yield only an estimate of the curve. Given that, we ground the evaluation on the notion of optimality, first introduced by AttackBench~\cite{cina2025attackbench}, to assess how effectively any given attack estimates it.

\myparagraph{Notation.} Let $f_{\vct \theta}: \mathcal{X} \rightarrow \mathbb{R}^C$ denote a classifier with parameters $\vct \theta$, where $\mathcal{X} = [0, 1]^d$ is the input domain and $\mathcal{Y} = \{1, \dots C\}$ is the set of class labels. 
Given an input $\vct x \in \mathcal{X}$, the model outputs a vector of logits $f_{\vct \theta}(\vct x) = [f_1(\vct x), \dots, f_C(\vct x)]$, and the predicted class is taken as $\arg\max_{c} f_c(\vct{x})$. 
Given the true label $y \in \mathcal{Y}$, an adversarial example $\vct{x}' = \vct{x} + \vct{\delta}$ satisfies $ \arg\max_{c} f_c(\vct{x}') \neq y, \text{with } \vct{x}' \in \mathcal{X}$. The perturbation size is measured by the $\ell_p$ norm $\|\vct{\delta}\|_p$, where $p \in \{0,1,2,\infty\}$.

\myparagraph{The robustness--perturbation curve.}
Standard evaluations report the \emph{robust accuracy} at a fixed budget $\varepsilon$, \ie, the fraction of samples whose prediction remains correct under any perturbation of size at most $\varepsilon$. Varying $\varepsilon$ traces the complete robustness--perturbation curve $\rho(\varepsilon; f_{\vct \theta})$, which characterizes robustness across a wide range of budgets rather than at a single operating point. Formally, let $\vct{\delta}^\star(\vct{x}, f_{\vct \theta})$ denote the minimal perturbation sufficient to induce misclassification for a given input $\vct{x}$. The ideal robustness--perturbation curve evaluated over a dataset $\mathcal{D}$ of input--label pairs $(\vct{x}, y) \in \DD$ is then defined as:
\begin{equation}
\label{eq:rob_curve_theoretical}
\rho^\star(\varepsilon; f_{\vct{\theta}}) =
\frac{1}{|\mathcal{D}|}\sum_{(\vct{x},y)\in\mathcal{D}}
\mathbb{I}[\,\|\vct{\delta}^\star(\vct{x}, f_{\vct\theta})\|_p > \varepsilon\,]\,,
\end{equation}
where $\mathbb{I}[\cdot]$ is the indicator function, \ie, $1$ if the condition is met and $0$ otherwise.

\subsection{Minimum-norm Adversarial Attacks}

To estimate the robustness--perturbation curve, and in particular the minimal distance $\|\vct{\delta}^\star(\vct{x}, f_{\vct\theta})\|_p$, we use minimum-norm attacks.

\myparagraph{Attack objective.}
For a model $f_{\vct \theta}$, the goal of a minimum-norm adversarial attack $a$ is to find the smallest perturbation that induces misclassification:
\begin{equation}
\label{eq:minnorm_loss}
\begin{aligned}
\vct{\delta}^\star \in \arg\min_{\vct{\delta}} \quad & \|\vct{\delta}\|_p \\
\text{s.t.} \quad & \arg\max_{c} f_c(\vct{x}+\vct{\delta}) \neq y, \\
& \vct{x}+\vct{\delta} \in \mathcal{X} .
\end{aligned}
\end{equation}
The constraint $\arg\max_c f_c(\vct{x}+\vct{\delta}) \neq y$ asks only for
\emph{some} misclassification, without fixing the resulting class; we call this
the \emph{untargeted} formulation. A \emph{targeted} run instead drives the
optimization toward a chosen class $t \neq y$. As any $\vct{\delta}$ that satisfies the constraint is still a valid solution, targeting serves only to guide the trajectory, and any misclassification it produces remains admissible.

This problem has no closed-form solution and is typically solved with iterative gradient-based solvers, which use the gradient of Eq. \eqref{eq:minnorm_loss} to refine the perturbation. 
Each returns a distance that upper-bounds the true minimal norm, $d(\vct{x}; f_{\vct\theta}) \ge \|\vct{\delta}^\star(\vct{x}, f_{\vct\theta})\|_p$. Substituting $d(\vct{x}; f_{\vct\theta})$ for $\|\vct{\delta}^\star(\vct{x}, f_{\vct\theta})\|_p$ in Eq.~\eqref{eq:rob_curve_theoretical} yields the empirical robustness--perturbation curve we compute as:
\begin{equation}
\label{eq:rob_curve}
\rho(\varepsilon; f_{\vct \theta}) = \frac{1}{|\DD|}\sum_{\vct{x}\in\DD}\mathbb{I}[d(\vct{x}; f_{\vct \theta})>\varepsilon] \,.
\end{equation}

\subsection{Empirical Attack Frontier and Attack Optimality}
\label{sec:attack-frontier}

Minimum-norm attack implementations, however, come with no measure of how effective they are: different solvers, or even different configurations of the same one, may return different solutions for the same sample. 
Comparing them, and knowing how close any comes to the smallest achievable perturbation, requires a quantitative notion of how close to optimality an attack is.
AttackBench~\cite{cina2025attackbench} provides a quantitative measure of optimality that is, by definition, relative to a predefined pool $\A$ of $N$ attacks.
Concretely, it first constructs the \emph{attack frontier}.
Given a model $f_{\vct\theta}$ and a fixed query budget $Q$ (\ie, the maximum allowed forward and backward passes through the model an attack can make), the frontier is defined as the per-sample minimum distance $d_{a^\star}(\vct{x}) = \min_{a\in\A} d_a(\vct{x})$ evaluated over each sample $\vct{x}\in\mathcal{D}$. 
Here, $a^\star$ represents the resulting per-sample optimal attack under the given budget constraint.
The curve $\rho_{a^\star}(\varepsilon; f_{\vct\theta})$ is the empirical robustness--perturbation curve relative to $\A$, the tightest estimate of $\rho^\star(\varepsilon; f_{\vct\theta})$, obtained using all attacks in the pool. 
The closer an attack's own curve lies to this frontier, the higher the \emph{optimality}. This is formalized as the \attackoptimality{} (\aoi{}) below.

Let $\rho_a(\varepsilon; f_{\vct\theta})$ denote the robustness--perturbation curve obtained by attack $a$ against model $f_{\vct\theta}$.  
The overall effectiveness of $a$ is quantified by the \emph{Area Under the Robustness Evaluation Curve} (AUREC):
\begin{equation}\label{eq:aurec}
    \mathrm{AUREC}_a(f_{\vct\theta})
    = \int_0^{\varepsilon_{\min}(f_{\vct\theta})} \rho_a(\varepsilon; f_{\vct\theta})\, d\varepsilon,
\end{equation}
where $\varepsilon_{\min}(f_{\vct\theta}) = \min\{\varepsilon \ge 0 : \rho_{a^\star}(\varepsilon; f_{\vct\theta}) = 0\}$ is the smallest perturbation budget at which the frontier reaches zero robust accuracy. Stronger attacks have smaller $\mathrm{AUREC}$, \ie, they succeed with smaller budgets.

The per-model \attackoptimality\ (\aoi) of attack $a$ is then obtained by comparison with $a^\star$:
\begin{equation}
\label{eq:aoi_local}
    \aoi(a, f_{\vct\theta})
    = \frac{\rho(0; f_{\vct\theta})\,\varepsilon_{\min}(f_{\vct\theta})
      - \mathrm{AUREC}_a(f_{\vct\theta})}
      {\rho(0; f_{\vct\theta})\,\varepsilon_{\min}(f_{\vct\theta})
      - \mathrm{AUREC}_{a^\star}(f_{\vct\theta})},
\end{equation}
where $\rho(0; f_{\vct\theta})$ is the clean accuracy of the model.  
By construction, $\aoi(a,f_{\vct\theta})\in[0,1]$, equals~1 only if $\rho_a(\varepsilon; f_{\vct\theta})=\rho_{a^\star}(\varepsilon; f_{\vct\theta})$ for all $\varepsilon\in[0,\varepsilon_{\min}]$, and otherwise measures the normalized area gap to the empirical frontier.

Since attack optimality is model-dependent, an attack may closely approximate the frontier for one defense but not for another. To summarize its performance across a set of models $\MM$ evaluated under the same $\ell_p$ norm, we define the \emph{global} \aoi{} (\gaoi{}) as the mean of the corresponding per-model values:
\begin{equation}
\label{eq:global_aoi}
    \gaoi(a)
    = \frac{1}{|\MM|} \sum_{f_{\vct\theta} \in \MM}
      \aoi(a, f_{\vct\theta}).
\end{equation}
By definition, $\gaoi(a)\in[0,1]$ and quantifies how closely $a$ approaches the empirical attack frontiers on average across models. 

In summary, given a pool of attacks, AttackBench compares their relative effectiveness by measuring the \gaoi{} of each attack independently over a set of models, using a fixed query budget.
This makes it possible to identify the attack that is, on average, the most effective for robustness evaluation. However, this approach has two important limitations:
\newline \noindent  \textit{(i) It evaluates attacks in isolation.} No individual attack is expected to approach the empirical attack frontier equally well across all samples and models. Different attacks may therefore be complementary: taking the best perturbation found by each attack on a per-sample basis can produce a
    tighter robustness curve than any individual attack.
\newline \noindent \textit{(ii) It assigns a fixed budget independently to every attack.}
    Because the query budget is split uniformly, each attack receives the same allocation regardless of its strength, so queries spent on weaker or redundant attacks are effectively wasted. An example can be found in our experiments (Table~\ref{tab:ensembles-overview}): some attacks are selected with 100-200 queries rather than using the maximum budget of 1000, as optimizing further rarely improves the perturbation found for a sample. A uniform allocation would instead spend the same budget on these members as on the dominant ones, wasting queries the frontier does not need.

To address these limitations, we extend the evaluation from individual attacks to attack ensembles and allocate a single total query budget across the attacks in the pool. 
The ensemble combines the complementary strengths of its members, thereby extending the search coverage without increasing the overall evaluation budget. 
This allows us to study both how the allocation of queries affects ensemble effectiveness and the budget required for the ensemble to approach the tightest empirical frontier attainable by the considered attack
pool. The central question is therefore how to distribute a fixed query budget across attacks so as to approximate this frontier as closely as possible. 

\section{Approximating the Attack Frontier with Attack Ensembles}\label{sec:methodology}

Our goal is to construct attack ensembles that approximate the attack frontier as closely as possible, under a given query budget a practitioner can control.
The resulting ensembles can then estimate the robustness--perturbation curve of any given model.
In the following, we present a greedy budget allocation approach that optimizes the \gaoi{} of the resulting ensemble.

\subsection{Problem formulation}\label{sec:problem}

Given a set of attacks $\A=\{a_1,\ldots,a_N\}$ and a set of models $\MM$, we aim to approximate the empirical attack frontier by combining a subset of the attacks in $\A$, under a fixed computational budget.
To make computation comparable across attacks, we measure cost in \emph{queries}, \ie, the sum of forward and backward passes through the target model.
This provides a simple and consistent unit of computation and allows different attacks to be evaluated under the same budget constraint.
We denote by $T$ the maximum number of queries any attack can spend, and by $Q$ the total number of queries available to the whole ensemble.
Ensemble construction thus reduces to deciding how many queries to assign to each attack in $\A$. This leads to a budget allocation formulation.

Under this view, an ensemble is fully specified by a vector of per-attack query budgets $\vct{q} = (q_a)_{a \in \A} \in \mathbb{N}^{N}$, where $N = |\A|$ and each entry $q_a$ specifies the number of queries allocated to attack $a$. 
The attacks actually included in the ensemble are those with a non-zero allocation:
\begin{equation}
\label{eq:qvec}
\ens = \{a \in \A : q_a > 0\} \subseteq \A.
\end{equation}
We can measure the optimality of the ensemble $\ens$ directly as a function of the query vector, $\gaoi(\vct{q})$, extending Eq.~\eqref{eq:global_aoi} to a complete budget allocation.

\myparagraph{Optimization Problem.}
The empirical attack frontier corresponds to the ideal ensemble $\ens_{max}$, obtained from the query vector $\vct{q}_{max}$, which sets $q_a = T$ for every attack $a \in \A$, \ie, runs the whole pool at the per-attack cap, so that $\ens_{max} = \A$. 
By construction, $\gaoi(\vct{q}_{max}) = 1$, though a smaller budget may suffice to reach it.
This regime defines a reference upper bound of performance but is not attainable under practical computational constraints.
We therefore seek for $\vct{q}^\star$, the best feasible approximation to $\vct{q}_{max}$ with a limited total query budget $Q \le N \cdot T$.
We formalize this as the following optimization problem:
\begin{equation}
\label{eq:fb-problem}
\vct{q}^\star \;\in\; \argmax_{\vct{q}\in\mathbb{N}^{N}}\ \gaoi(\vct{q})
\quad\text{s.t.}\quad
\underbrace{\sum_{a\in\A} q_a \le Q}_{\text{total budget}},\ \
\underbrace{q_a\le T\ \forall a}_{\text{per-attack cap}},\ \
\vct{q}\in\mathcal{C}.
\end{equation}
where $\mathcal{C} \subset \mathbb{N}^{N}$ encodes additional structural rules to include the same attack with multiple variants, discussed in Sect.~\ref{sec:exp-setup}.
We denote by $\ens^\star$ the ensemble induced by $\vct{q}^\star$ that attains the highest \gaoi{} achievable given the available query budget $Q$. 
Since $\vct{q}^\star$ cannot be computed exactly, in Section~\ref{sec:greedy-alg} we present a greedy algorithm that returns an ensemble $\ensopt$ that approximates $\ens^\star$.

\subsection{Greedy algorithm}\label{sec:greedy-alg}

Problem~\eqref{eq:fb-problem} is a discrete budgeted maximization whose search space grows exponentially with the pool size, therefore cannot be solved exhaustively.
To find a suitable solution, we propose the greedy procedure reported in Algorithm~\ref{alg:greedy}. 

We start from the empty allocation $\vct{q} \equiv \vct{0}$ (Line~\ref{algo:init}), i.e., the induced ensemble is empty.
For a given attack $a \in \A$, a candidate update raises its budget to a strictly larger value $\hat q \in \mathcal{S}$, \ie, it sets $q_a \equiv \hat q$ while leaving every other entry of $\vct{q}$ unchanged. 
We denote by $\vct{q}'$ the vector resulting from this update.
We consider only updates for which the resulting query vector satisfies the constraints in Eq.~\eqref{eq:fb-problem}.
Given the current query allocation $\vct q$, the \textit{feasible} updates are:
\begin{equation}\label{eq:moveset}
\mathcal{N}(\vct q)=\Bigl\{\,(a,\hat q)\in\A\times\mathcal{S}\ :\ \hat q>q_a,\ \ \vct{q}'\in\mathcal{C},\ \ \textstyle\sum_{b\in\A}q_b-q_a+\hat q\le Q\Bigr\},
\end{equation}
Each feasible update is scored by its \gaoi gain, scaled by the added queries, \ie,
\begin{equation}
\label{eq:density}
g_{a,\hat q}\;=\;\frac{\gaoi(\vct{q}')-\gaoi(\vct{q})}{\hat q - q_a}\,.
\end{equation}
The normalization promotes query-efficient improvements over raw gains.

We iteratively select the update in $\mathcal{N}(\vct{q})$ with the largest $g_{a,\hat q}$ (Line~\ref{algo:argmax}) and apply it (Line~\ref{algo:update}), achieving a heuristic algorithm that approximates the frontier within the query budget $Q$.
The algorithm stops (Line~\ref{algo:while}) when it reaches the maximum allocated budget $Q$ (\ie, when the feasible set of candidate updates $\mathcal{N}(\vct{q})$ is empty), or when there is no update that improves the overall \gaoi{} of the ensemble. Finally, the algorithm returns the query vector $\widehat{\vct{q}}$ it has constructed (Line~\ref{algo:return}), whose induced ensemble is $\ensopt$.

\begin{algorithm}[t]
\LinesNumbered
\SetKwInOut{Input}{Input}
\SetKwInOut{Output}{Output}
\SetKwComment{Comment}{$\triangleright$\ }{}
\DontPrintSemicolon
\caption{Greedy Ensemble Construction.}
\label{alg:greedy}
\small
\Input{attack pool $\A$; models $\MM$; query budget $Q$.}
\Output{query vector $\widehat{\vct{q}}$ maximizing $\gaoi$ subject to the budget (Eq.~\ref{eq:fb-problem}), whose induced ensemble is $\ensopt$.}

$\vct{q} \equiv \vct{0}$ \Comment*[r]{empty ensemble} \label{algo:init}
\While{$\exists\,(a,\hat q)\in\mathcal{N}(\vct{q})$ \textbf{with} $g_{a,\hat q}>0$}{ \label{algo:while}
  $(a^\star,\hat q^\star)\gets\displaystyle\argmax_{(a,\hat q)\in\mathcal{N}(\vct{q})}\ g_{a,\hat q}$
  \Comment*[r]{best gain per added query} \label{algo:argmax}
  $q_{a^\star} \equiv \hat q^\star$ \Comment*[r]{update queries} \label{algo:update}
}
\textbf{return}\ $\widehat{\vct{q}}\gets\vct{q}$ \label{algo:return}
\end{algorithm}

The ensemble $\ensopt$ obtained from Algorithm~\ref{alg:greedy} can be used to evaluate the robustness of any given model, yielding its robustness--perturbation curve. 
The allocation procedure is agnostic to the attacks in the pool. To strengthen the pool used in our experiments, we add \APGDmin{}.

\subsection{Extending the Pool: Minimum-Norm Auto-Projected Gradient Descent}\label{sec:apgdmin}

Auto-Projected Gradient Descent (APGD)~\cite{croce2020reliable,croce2022sparse} is one of the best attacks for $\ell_1$ and $\ell_\infty$, despite having been designed for fixed-radius evaluation. 
This motivates \APGDmin{}, our native minimum-norm adaptation of APGD for $\ell_1$, $\ell_2$, and $\ell_\infty$, which searches for small perturbations in one run rather than repeatedly invoking a point-wise attack at different radii.

\APGDmin{} retains APGD's projected-gradient updates but adapts the radius $\varepsilon_t$ per sample. It initializes the radius from a first-order boundary estimate using the dual-norm gradient $\|\vct g_t\|_q$. Following FMN~\cite{pintor2021fast}, it tracks the best perturbation $\vct\delta^\star$, shrinks the radius toward $\|\vct\delta^\star\|_p$ after a successful iterate, and expands it otherwise. The update rate uses cosine decay, the step size scales with $\varepsilon_{t+1}/\varepsilon_t$, and checkpoints halve the step size and restart from the best adversarial example. For $\ell_2$ and $\ell_\infty$, these are the only changes to the original APGD. For $\ell_1$, \APGDmin{} retains the sparse-sign direction, no-momentum update, exact projection onto $\mathcal{B}_1(\varepsilon_t)\cap[0,1]^d$, and sparsity-ratio checkpoint of $\ell_1$-APGD.
The implementation can be found in our released code (see Section~\ref{sec:experiments}).

\section{Ranking Defenses: the Defense Optimality Index}
\label{sec:doi}
Section~\ref{sec:methodology} showed how to construct, for a given query budget, an attack ensemble that best approximates the attack frontier.
We now use this ensemble to evaluate and rank a given set of defenses.
To do so, we introduce the \emph{defense frontier} and the \emph{Defense Optimality Index} (\doi{}), the defense-side counterparts of the attack frontier and the \aoi{} introduced in Section~\ref{sec:attack-frontier}.
We detail below how we build the frontier and compute the \doi{}, which scores each defense by how close its robustness--perturbation curve is to the defense frontier.

\myparagraph{Defense Frontier.}
Let $d_{\ens}(\vct{x}; f_{\vct\theta})$ be the distance per-sample found by the attack ensemble $\ens$ on a model $f_{\vct\theta} \in \mathcal M$. 
Retaining the maximum of these minima across all models defines the \textit{defense frontier}, \ie, an oracle model that misclassifies each sample with the largest minimum-norm perturbation:
\begin{equation}
    d_{\mathrm{def}}^\star(\vct{x}) = \max_{f_{\vct\theta} \in \MM}
        d_{\ens}(\vct{x}; f_{\vct\theta}).
\end{equation}
The corresponding robustness--perturbation curve $\rho_{\mathrm{def}}^\star(\varepsilon)$ follows from Eq.~\eqref{eq:rob_curve}, and its area, $\mathrm{AUREC}_{\mathrm{def}}^\star$, from Eq.~\eqref{eq:aurec}.
The integration in this case runs up to $\varepsilon_{\mathrm{def}} = \min\{\varepsilon \ge 0 : \rho_{\mathrm{def}}^\star(\varepsilon) = 0\}$, the smallest budget at which the defense frontier reaches zero. The same upper limit is used for every model in $\MM$, making the areas directly comparable.

\begin{table*}[!t]
\centering
\caption{\textbf{Benchmark model pool.} CIFAR-10 (C1--C20) and ImageNet (I1--I10) defenses used in our experiments. The \emph{Norm} columns denote the perturbation norm $\ell_p$ each model was adversarially trained on; BS is the batch size.}
\label{tab:model_pool}
\resizebox{\textwidth}{!}{%
\setlength{\tabcolsep}{4pt}
\begin{tabular}{@{}clcc@{\hskip 14pt}clcc@{\hskip 14pt}clcc@{}}
\toprule
\multicolumn{8}{c}{\textit{CIFAR-10}} & \multicolumn{4}{c}{\textit{ImageNet}} \\
\cmidrule(r){1-8}\cmidrule(l){9-12}
\textbf{ID} & \textbf{Model} & \textbf{Norm} & \textbf{BS} &
\textbf{ID} & \textbf{Model} & \textbf{Norm} & \textbf{BS} &
\textbf{ID} & \textbf{Model} & \textbf{Norm} & \textbf{BS} \\
\midrule
C1  & Addepalli~\cite{addepalli2022scaling}   & $\infty$ & 500 & C11 & Rade~\cite{rade2021helper}             & $2$         & 500 & I1  & Chen~\cite{chen2024data}                           & $\infty$ & 100 \\
C2  & Augustin~\cite{augustin2020adversarial} & $2$      & 500 & C12 & Rebuffi~\cite{rebuffi2021fixing}       & $\infty$    & 500 & I2  & Debenedetti~\cite{debenedetti2023light}            & $\infty$ & 100 \\
C3  & Chen~\cite{chen2024data}                & $\infty$ & 500 & C13 & Sehwag~\cite{sehwag2021robust}         & $2$         & 500 & I3  & Rodriguez-Munoz~\cite{rodriguez2024characterizing} & $\infty$ & 100 \\
C4  & Cui~\cite{cui2024decoupled}             & $\infty$ & 500 & C14 & Stutz~\cite{stutz2020confidence}       & $\infty$    & 500 & I4  & Singh~\cite{singh2023revisiting}                   & $\infty$ & 100 \\
C5  & Engstrom                                 & $2$      & 500 & C15 & Zhong~\cite{zhong2024towards} (PGD)    & $0_{k=120}$ & 500 & I5  & Wong~\cite{wong2020fast}                           & $\infty$ & 200 \\
C6  & Gowal~\cite{gowal2021improving}         & $\infty$ & 250 & C16 & Zhong~\cite{zhong2024towards} (TRADES) & $0_{k=120}$ & 500 & I6  & Engstrom                                            & $\infty$ & 200 \\
C7  & Gowal~\cite{gowal2021improving}         & $\infty$ & 500 & C17 & Pang~\cite{pang2022robustness}         & $\infty$    & 250 & I7  & Salman~\cite{salman2020adversarially}              & $\infty$ & 200 \\
C8  & Jiang~\cite{jiang2023towards}           & $1$      & 500 & C18 & Engstrom                                & $\infty$    & 500 & I8  & Salman~\cite{salman2020adversarially}              & $\infty$ & 200 \\
C9  & Maini~\cite{maini2020adversarial}       & $1{,}2{,}\infty$ & 500 & C19 & Wang~\cite{wang2023better}     & $\infty$    & 500 & I9  & Singh~\cite{singh2023revisiting}                   & $\infty$ & 100 \\
C10 & Rade~\cite{rade2021helper}              & $\infty$ & 500 & C20 & Xu~\cite{xu2023exploring}              & $\infty$    & 500 & I10 & Singh~\cite{singh2023revisiting}                   & $\infty$ & 250 \\
\bottomrule
\end{tabular}}
\vspace{2pt}\\
\end{table*}

\begin{table}[!t]
\centering
\caption{\textbf{Minimum-Norm Attack Pool.} Minimum-norm solvers for our evaluation ensembles, and their supported $\ell_p$ perturbation norms. \emph{Restarts} indicates the number of random $\ell_p$-ball initializations ($-$ if unsupported). All solvers natively support targeted searches.}
\label{tab:attack_pool}
\resizebox{\linewidth}{!}{%
\setlength{\tabcolsep}{5pt}
\begin{tabular}{@{}l cccc c@{}}
\toprule
& \multicolumn{4}{c}{\textbf{Norms}} & \\
\cmidrule(lr){2-5}
\textbf{Solver} & $\ell_0$ & $\ell_1$ & $\ell_2$ & $\ell_\infty$ & \textbf{Restarts} \\
\midrule
Primal-Dual Proximal Gradient Descent (\pdpgd{})~\cite{matyasko2021pdpgd} & \cmark & \cmark & \cmark & \cmark & 5 \\
Primal-Dual Gradient Descent (\pdgd{})~\cite{matyasko2021pdpgd}           &        &        & \cmark &        & 5 \\
Fast Minimum-Norm (\fmn{})~\cite{pintor2021fast}                          & \cmark & \cmark & \cmark & \cmark & 5 \\
Auto-Projected Gradient Descent (\APGDmin{}) \textit{(Ours)}              &        & \cmark & \cmark & \cmark & 5 \\
Augmented Lagrangian Method for Adversarial (\alma{})~\cite{rony2020augmented} &    & \cmark & \cmark &        & --- \\
Decoupled Direction and Norm (\ddn{})~\cite{rony2019decoupling}           &        &        & \cmark &        & --- \\
$\sigma$-Zero (\sigmazero{})~\cite{cina2025sigmazero}                     & \cmark &        &        &        & 5 \\
\bottomrule
\end{tabular}
}
\end{table}

\myparagraph{Defense optimality.}
Let $\mathrm{AUREC}_{\ens}(f_{\vct\theta})$ be the area of model $f_{\vct\theta}$'s robustness--perturbation curve computed with  the ensemble $\ens$. We define the \defenseoptimality{} (\doi) as:
\begin{equation}
\label{eq:doi}
    \doi(f_{\vct\theta})
    = \frac{\mathrm{AUREC}_{\ens}(f_{\vct\theta})}
            {\mathrm{AUREC}_{\mathrm{def}}^\star} \,.
\end{equation}
By construction, $\doi(f_{\vct\theta})\in[0,1]$ and equals~1 only if $\rho_{\ens}(\varepsilon; f_{\vct\theta})=\rho_{\mathrm{def}}^\star(\varepsilon)$ for all $\varepsilon\in[0,\varepsilon_{\mathrm{def}}]$. Lower values indicate
a larger empirical gap from the best achievable robustness within $\MM$. 
As for \aoi, the metric is pool-relative and empirical. Its values depend on the perturbation norm, the strength of the ensemble used to compute the distances, and the diversity of the models in $\MM$.

Taken together, Sections~\ref{sec:methodology} and~\ref{sec:doi} complete our framework.
Given a query budget, the greedy algorithm of Section~\ref{sec:methodology} constructs an attack ensemble that approximates the attack frontier.
As the budget increases, the algorithm yields a nested family of ensembles, each an incremental extension of the previous one.
In Section~\ref{sec:doi} we use these ensembles to compute the \doi{} of a set of defenses, and evaluate their gap to the defense frontier.
In Section~\ref{sec:experiments}, we validate this framework empirically. 
We construct ensembles increasing query budget tiers and show that they closely approximate the attack frontier, with the approximation improving consistently as the budget grows across tiers.
We then use these ensembles to compute the \doi{} of a large pool of defenses, showing how this $\varepsilon$-independent ranking overcomes the brittleness of single-budget evaluation.

\section{Experiments}
\label{sec:experiments}
We validate our framework empirically in two steps: we first show that our ensembles closely approximate the attack frontier at a controllable query cost, and then use them to compute the \doi{} and demonstrate its stability against the brittleness of single-$\varepsilon$ evaluation. We run these experiments on a single NVIDIA H100 NVL (94\,GB). Our code and benchmark are accessible at \url{adversarial-frontier.github.io}.

\subsection{Setup}
\label{sec:exp-setup}

\myparagraph{Datasets and models.}
We evaluate our approach on \emph{CIFAR-10} and \emph{ImageNet} using a fixed $1\,000$-sample validation subset for each model--norm pair. We evaluate $30$ robust models, $20$ on CIFAR-10 and $10$ on ImageNet, drawn from RobustBench and their original sources. We label CIFAR-10 models as C1--C20 and ImageNet ones as I1--I10, as listed in Table~\ref{tab:model_pool}\footnote{Models C5, C18, and I6 are sourced from \url{https://github.com/madrylab/robustness}. For I10, the batch size is reduced to $100$ for standard AutoAttack and for the ablation grid on $\ell_1$ and $\ell_2$.}. We split these $30$ defenses into two partitions, \textit{train} and \textit{held-out}, stratified across datasets (seed $0$): $17$ training and $13$ held-out models ($12/8$ on CIFAR-10, $5/5$ on ImageNet). We compute the ensembles on the \textit{train} pool, and test them on the the unseen models of the \textit{held-out} set.

\myparagraph{Attack pool.} Table~\ref{tab:attack_pool} summarizes the pool of minimum-norm solvers\footnote{Attack implementations are adopted from \url{https://github.com/jeromerony/adversarial-library}.} we used across the four $\ell_p$ norms, and expanded by introducing our novel minimum-norm adaptation of Auto-Projected Gradient Descent (\APGDmin{}), we described in Section~\ref{sec:apgdmin}. All baseline solvers are executed with their default hyperparameters, The only exception is \pdpgd{} in $\ell_0$, where we replace the standard hard-thresholding with the closed-form $\ell_{2/3}$ concave surrogate proximal operator to improve sparsity. Finally, each solver's untargeted execution (seed 42) is augmented with two search variants. First, we execute \emph{targeted attacks} directed at the top-9 highest-confidence incorrect classes, yet still looking for an untargeted misclassification. Second, we employ \emph{random restarts} by evaluating 5 independent initializations sampled from the $\ell_p$ ball (seeds 43 to 47).

\begin{table}[!t]
\centering
\scriptsize
\setlength{\tabcolsep}{2.5pt}
\renewcommand{\arraystretch}{1.05}
\caption{\textbf{Deployable attack ensembles produced by the greedy construction.} Indented \emph{Restarts} and \emph{Targeted} rows report cumulative queries for the attack above; an en-dash denotes an inactive group. Details: \protect\url{https://github.com/adversarial-frontier.github.io}.}
\label{tab:ensembles-overview}
\begin{tabular*}{\linewidth}{@{\extracolsep{\fill}}lrrr@{\hspace{0.75em}}lrrr@{}}
\toprule
\textbf{Configuration} & $4$k & $8$k & $12$k &
\textbf{Configuration} & $4$k & $8$k & $12$k \\
\midrule
\multicolumn{4}{@{}c}{\textbf{$\ell_0$}} &
\multicolumn{4}{c@{}}{\textbf{$\ell_1$}} \\
\cmidrule(r){1-4} \cmidrule(l){5-8}

\sigmazero{}                & 800 & 900 & 900 & \APGDmin-\CE{}                & 1000 & 1000 & 1000 \\
\quad\emph{Restarts}       & 2400 & 3400 & 4400 & \quad\emph{Restarts}           & 900 & 2000 & 4000 \\
\quad\emph{Targeted}       & -- & 1600 & 3200 & \APGDmin-\DLR{}               & 1000 & 1000 & 1000 \\
\pdpgd{}                    & 600 & 700 & 700 & \quad\emph{Restarts}           & -- & 2000 & 3000 \\
\quad\emph{Restarts}       & -- & -- & 500 & \pdpgd{}                       & 400 & 600 & 600 \\
\quad\emph{Targeted}       & -- & 1200 & 2000 & \quad\emph{Restarts}           & 400 & 1100 & 2100 \\
\fmn{}                      & 200 & 200 & 200 & \fmn{}                         & 100 & 100 & 100 \\
\quad\emph{Restarts}       & -- & -- & 100 & \quad\emph{Restarts}           & 200 & 200 & 200 \\

\midrule
\multicolumn{4}{@{}c}{\textbf{$\ell_2$}} &
\multicolumn{4}{c@{}}{\textbf{$\ell_\infty$}} \\
\cmidrule(r){1-4} \cmidrule(l){5-8}

\alma{}                     & 1000 & 1000 & 1000 & \fmn{}                         & 800 & 1000 & 1000 \\
\quad\emph{Targeted}       & -- & -- & 2000 &                                &     &      &      \\
\pdgd{}                     & 400 & 500 & 1000 & \pdpgd{}                       & -- & -- & 800 \\
\quad\emph{Restarts}       & 1000 & 1600 & 2600 & \quad\emph{Restarts}           & -- & -- & 2700 \\
\quad\emph{Targeted}       & -- & 2200 & 2500 & \APGDmin-\CE{}                & 700 & 700 & 700 \\
\APGDmin-\DLR{}            & 400 & 700 & 700 & \quad\emph{Restarts}           & 100 & 2100 & 2400 \\
\quad\emph{Restarts}       & 700 & 1100 & 1100 & \quad\emph{Targeted}           & -- & -- & 100 \\
\APGDmin-\CE{}             & 500 & 600 & 600 & \APGDmin-\DLR{}               & 700 & 700 & 700 \\
\quad\emph{Restarts}       & -- & 300 & 300 & \quad\emph{Restarts}           & 1700 & 3500 & 3600 \\
\quad\emph{Targeted}       & -- & -- & 200 &                                &     &      &      \\

\bottomrule
\end{tabular*}
\end{table}

\begin{figure*}[!t]
    \centering
    \includegraphics[width=\linewidth]{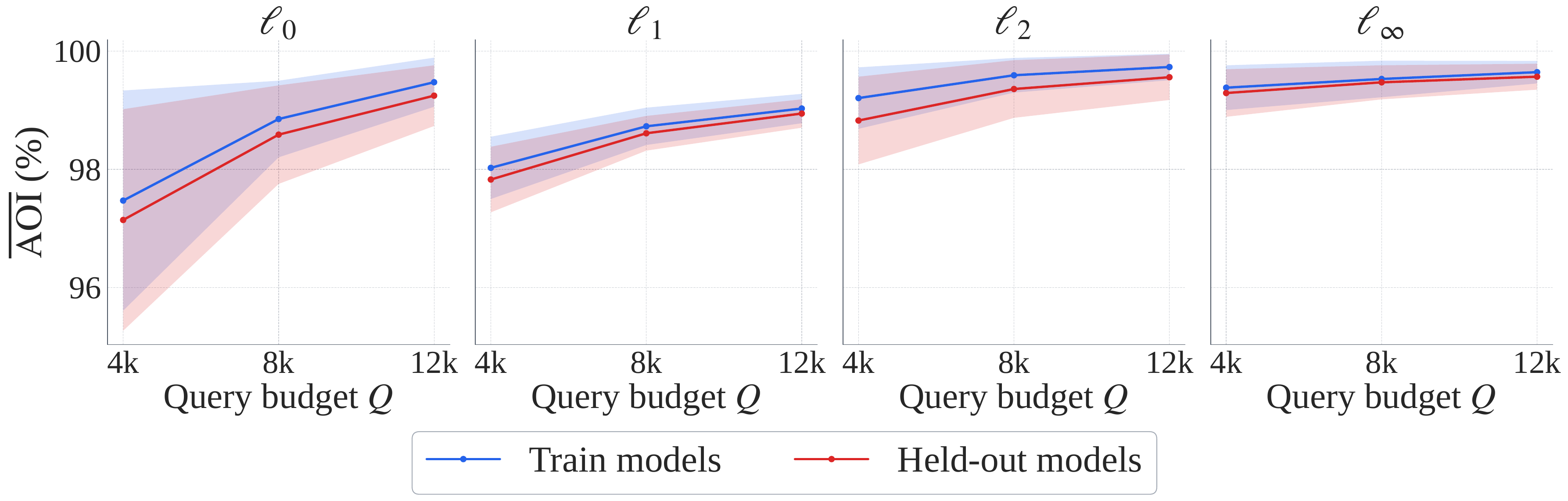}
    \caption{%
        \textbf{Generalization of attack optimality across query budgets.} 
        This plot shows \gaoi{} as a function of the query budget $Q$ across the four evaluated perturbation norms. 
        Blue lines denote the \attackoptimality{} on average across the set of $17$ training models, while red lines denote the same measure on average across $13$ held-out models.
        The \gaoi{} monotonically improves as the budget increases from $4\text{k}$ to $12\text{k}$ queries. 
        Crucially, the narrow gap between the train and held-out curves demonstrates that our greedy budget allocation generalizes effectively to unseen defenses without overfitting.
    }
    \label{fig:aoi_generalization}
\end{figure*}

We measure attack cost in \emph{queries}, defined as the total number of forward and backward passes through the model. Each attack is executed for $500$ optimization steps (equivalent to a maximum per-attack budget of $T = 1000$ queries), recording the perturbation distance every $10$ queries. Recall from Section~\ref{sec:problem} that the feasible set $\mathcal{C}\subset\mathbb{N}^N$ restricts the admissible query allocation vectors $\vct{q}$. Here, $\mathcal{C}$ enforces two structural dependencies tied to the pool expansion: (i) a targeted run against the $k$-th most likely incorrect class cannot receive more queries than the run against the $(k{-}1)$-th class, and (ii) a random restart may only receive queries if its corresponding untargeted base run is active. 

Candidate budget levels are drawn from $\mathcal{S}=\{100,200,\dots,1000\}$, defining a grid of $J=10$ evenly spaced checkpoints up to the cap $T$. Because our greedy strategy monotonically increases budget allocations, we record the resulting ensembles at three distinct query tiers, \ie, $Q\in\{4,8,12\}\mathrm{k}$ queries. This yields a nested hierarchy of ensembles $\ens_{4\mathrm{k}}\subseteq\ens_{8\mathrm{k}}\subseteq\ens_{12\mathrm{k}}$. A user can then pick the tier matching their compute budget and later extend it by resuming the same run.

\myparagraph{Metrics and baselines.}
For each model and query tier, we report the per-model \aoi{} and robust accuracy at standard AA reference budgets $\varepsilon_{\mathrm{ref}}$. For CIFAR-10, these are set to $12$, $0.5$, and $8/255$ for $\ell_1$, $\ell_2$, and $\ell_\infty$, respectively; for ImageNet, they are $40$, $3.0$, and $4/255$.
Finally, we use the $\ens_{12\mathrm{k}}$ ensemble to rank all $30$ models by computing the \doi{}.

We compare our approach against AA~\cite{croce2020reliable}. Since AA reports robust accuracy at a single fixed $\varepsilon$ and returns no per-sample minimum-norm distance, it provides one evaluation point rather than a complete curve, and cannot be scored by \aoi{} directly. We thus approximate its curve by evaluating AA over a per-model grid of $\varepsilon$ values, derived from our attack frontiers as $9$ distance percentiles plus the reference budget $\varepsilon_{\mathrm{ref}}$. We then linearly interpolate these points to reconstruct AA's robustness--perturbation curve and compute its \aoi{}.

\subsection{Results}\label{sec:results}
We first validate the three ensembles on $13$ held-out defenses, showing that their \aoi{} generalizes to unseen models and matches or exceeds AA's, both on the complete robustness--perturbation curve and at the single reference budget $\varepsilon_{\mathrm{ref}}$. Once validated, $\ens_{12\mathrm{k}}$ becomes our evaluation tool: we use it on all $30$ defenses to show how the \doi{} resolves the limitations of fixed-$\varepsilon$ evaluation, assessing every defense across all four perturbation norms, even beyond the one used for its own adversarial training, and beyond RobustBench's coverage.

\myparagraph{Ensemble composition and generalization.}
Table~\ref{tab:ensembles-overview} summarizes the composition at each query tier. Base rows report the queries assigned to one run, whereas \emph{Restarts} and \emph{Targeted} rows report the total across all runs in that group; per-run details are available on our benchmark website. Composition varies by norm and budget. The $\ell_0$ ensembles center on \sigmazero{}, with smaller contributions from \pdpgd{} and \fmn{}. Under $\ell_1$, they rely mainly on the base and restart variants of both \APGDmin{} losses. The broader $\ell_2$ ensembles are led by \pdgd{} restarts and targeted runs, complemented by \alma{} and \APGDmin{}. Under $\ell_\infty$, the allocation concentrates on \APGDmin{} and \pdpgd{} restarts, with a single \fmn{} base run. The learned allocations also generalize beyond the $17$ training models. Figure~\ref{fig:aoi_generalization} compares the \gaoi{} on these models with that on the $13$ held-out models from $4$k to $12$k queries. The curves track closely across all norms and tiers, with a maximum gap below $0.4$ percentage points, indicating no substantial degradation on unseen models.

\myparagraph{Optimality analysis.}
Table~\ref{tab:ours-vs-aa-tiers} shows that, at $Q=12$k, our ensembles match or exceed AA's \aoi{} on all $13$ held-out models on $\ell_1$, on $12$ out of $13$ on $\ell_2$, and on $10$ out of $13$ on $\ell_\infty$; their mean \aoi{} is higher and its variation lower on all three norms. The largest improvement is on $\ell_2$, where the $4$k ensemble already matches or exceeds AA on $12$ of $13$ held-out models. Because \aoi{} evaluates complete robustness--perturbation curves against the empirical attack frontier, these results demonstrate an improved frontier approximation rather than only robust accuracy at a selected perturbation radius.

\begin{table*}[!t]
\centering
\scriptsize
\setlength{\tabcolsep}{4pt}
\caption{\textbf{Our greedy ensembles} vs.\ AA: \aoi (\%) per model as the query budget grows ($4$k$\to$$12$k), train and held-out models. \hlbest{Green}: our \aoi matches or exceeds AutoAttack.} 
\label{tab:ours-vs-aa-tiers}
\resizebox{\textwidth}{!}{%
\begin{tabular}{@{}c@{\hspace{8pt}}c@{\hspace{8pt}}c@{}}
\begin{tabular}[t]{@{}l c ccc@{}}
\multicolumn{5}{c}{\textbf{$\ell_1$}} \\
\toprule
ID & AA & $\ens_{4\mathrm{k}}$ & $\ens_{8\mathrm{k}}$ & $\ens_{12\mathrm{k}}$ \\
\midrule
\multicolumn{5}{@{}l}{\itshape Train} \\
C1 & 96.54 & \cellcolor{bestbg}96.97 & \cellcolor{bestbg}97.95 & \cellcolor{bestbg}98.54 \\
C2 & 98.25 & \cellcolor{bestbg}98.42 & \cellcolor{bestbg}99.07 & \cellcolor{bestbg}99.39 \\
C5 & 99.32 & 97.79 & 98.56 & 98.86 \\
C7 & 98.50 & 98.16 & \cellcolor{bestbg}98.83 & \cellcolor{bestbg}99.07 \\
C8 & 98.82 & 97.96 & 98.59 & \cellcolor{bestbg}98.82 \\
C9 & 97.61 & \cellcolor{bestbg}98.28 & \cellcolor{bestbg}98.76 & \cellcolor{bestbg}99.08 \\
C11 & 98.84 & 97.96 & \cellcolor{bestbg}98.91 & \cellcolor{bestbg}99.16 \\
C13 & 98.88 & 97.58 & 98.51 & 98.82 \\
C14 & 93.61 & \cellcolor{bestbg}97.53 & \cellcolor{bestbg}98.72 & \cellcolor{bestbg}99.20 \\
C15 & 98.97 & 98.42 & 98.94 & \cellcolor{bestbg}99.21 \\
C17 & 97.33 & \cellcolor{bestbg}97.83 & \cellcolor{bestbg}98.66 & \cellcolor{bestbg}99.04 \\
C20 & 97.45 & \cellcolor{bestbg}97.45 & \cellcolor{bestbg}98.54 & \cellcolor{bestbg}98.89 \\
I1 & 98.39 & \cellcolor{bestbg}98.74 & \cellcolor{bestbg}99.12 & \cellcolor{bestbg}99.32 \\
I2 & 98.32 & 98.18 & \cellcolor{bestbg}98.68 & \cellcolor{bestbg}98.89 \\
I4 & 97.68 & 97.56 & \cellcolor{bestbg}98.34 & \cellcolor{bestbg}98.66 \\
I5 & 99.19 & 99.02 & \cellcolor{bestbg}99.28 & \cellcolor{bestbg}99.42 \\
I7 & 98.59 & 98.58 & \cellcolor{bestbg}98.91 & \cellcolor{bestbg}99.11 \\
\midrule
\multicolumn{5}{@{}l}{\itshape Held-out} \\
C3 & 97.54 & \cellcolor{bestbg}97.98 & \cellcolor{bestbg}98.78 & \cellcolor{bestbg}99.19 \\
C4 & 97.57 & \cellcolor{bestbg}97.83 & \cellcolor{bestbg}98.68 & \cellcolor{bestbg}99.07 \\
C6 & 97.33 & \cellcolor{bestbg}97.95 & \cellcolor{bestbg}98.70 & \cellcolor{bestbg}99.00 \\
C10 & 97.53 & 96.58 & \cellcolor{bestbg}98.20 & \cellcolor{bestbg}98.77 \\
C12 & 97.76 & \cellcolor{bestbg}97.80 & \cellcolor{bestbg}98.64 & \cellcolor{bestbg}98.97 \\
C16 & 99.22 & 98.52 & 99.01 & \cellcolor{bestbg}99.22 \\
C18 & 98.17 & 97.48 & \cellcolor{bestbg}98.43 & \cellcolor{bestbg}98.78 \\
C19 & 97.41 & \cellcolor{bestbg}97.59 & \cellcolor{bestbg}98.53 & \cellcolor{bestbg}98.90 \\
I3 & 96.26 & \cellcolor{bestbg}97.08 & \cellcolor{bestbg}97.95 & \cellcolor{bestbg}98.31 \\
I6 & 98.36 & \cellcolor{bestbg}98.40 & \cellcolor{bestbg}98.85 & \cellcolor{bestbg}99.04 \\
I8 & 98.38 & 98.24 & \cellcolor{bestbg}98.71 & \cellcolor{bestbg}99.04 \\
I9 & 97.93 & 97.87 & \cellcolor{bestbg}98.51 & \cellcolor{bestbg}98.83 \\
I10 & 98.31 & \cellcolor{bestbg}98.44 & \cellcolor{bestbg}98.94 & \cellcolor{bestbg}99.15 \\
\bottomrule
\end{tabular}
&
\begin{tabular}[t]{@{}l c ccc@{}}
\multicolumn{5}{c}{\textbf{$\ell_2$}} \\
\toprule
ID & AA & $\ens_{4\mathrm{k}}$ & $\ens_{8\mathrm{k}}$ & $\ens_{12\mathrm{k}}$ \\
\midrule
\multicolumn{5}{@{}l}{\itshape Train} \\
C1 & 97.65 & \cellcolor{bestbg}99.07 & \cellcolor{bestbg}99.49 & \cellcolor{bestbg}99.72 \\
C2 & 98.79 & \cellcolor{bestbg}99.49 & \cellcolor{bestbg}99.71 & \cellcolor{bestbg}99.83 \\
C5 & 99.32 & \cellcolor{bestbg}99.71 & \cellcolor{bestbg}99.87 & \cellcolor{bestbg}99.92 \\
C7 & 98.24 & \cellcolor{bestbg}99.28 & \cellcolor{bestbg}99.66 & \cellcolor{bestbg}99.79 \\
C8 & 99.75 & 99.70 & \cellcolor{bestbg}99.92 & \cellcolor{bestbg}99.95 \\
C9 & 99.12 & \cellcolor{bestbg}99.53 & \cellcolor{bestbg}99.72 & \cellcolor{bestbg}99.86 \\
C11 & 99.76 & 99.74 & \cellcolor{bestbg}99.88 & \cellcolor{bestbg}99.91 \\
C13 & 99.25 & \cellcolor{bestbg}99.52 & \cellcolor{bestbg}99.81 & \cellcolor{bestbg}99.87 \\
C14 & 97.67 & \cellcolor{bestbg}98.60 & \cellcolor{bestbg}99.24 & \cellcolor{bestbg}99.41 \\
C15 & 99.86 & 99.74 & 99.82 & \cellcolor{bestbg}99.94 \\
C17 & 98.11 & \cellcolor{bestbg}99.30 & \cellcolor{bestbg}99.68 & \cellcolor{bestbg}99.81 \\
C20 & 98.15 & \cellcolor{bestbg}99.08 & \cellcolor{bestbg}99.50 & \cellcolor{bestbg}99.65 \\
I1 & 97.15 & \cellcolor{bestbg}99.04 & \cellcolor{bestbg}99.56 & \cellcolor{bestbg}99.72 \\
I2 & 97.77 & \cellcolor{bestbg}99.07 & \cellcolor{bestbg}99.52 & \cellcolor{bestbg}99.66 \\
I4 & 95.49 & \cellcolor{bestbg}97.68 & \cellcolor{bestbg}98.74 & \cellcolor{bestbg}99.07 \\
I5 & 97.34 & \cellcolor{bestbg}98.79 & \cellcolor{bestbg}99.30 & \cellcolor{bestbg}99.56 \\
I7 & 97.57 & \cellcolor{bestbg}99.15 & \cellcolor{bestbg}99.63 & \cellcolor{bestbg}99.75 \\
\midrule
\multicolumn{5}{@{}l}{\itshape Held-out} \\
C3 & 98.06 & \cellcolor{bestbg}99.26 & \cellcolor{bestbg}99.63 & \cellcolor{bestbg}99.78 \\
C4 & 97.83 & \cellcolor{bestbg}99.03 & \cellcolor{bestbg}99.43 & \cellcolor{bestbg}99.62 \\
C6 & 97.19 & \cellcolor{bestbg}98.93 & \cellcolor{bestbg}99.49 & \cellcolor{bestbg}99.69 \\
C10 & 96.61 & \cellcolor{bestbg}99.17 & \cellcolor{bestbg}99.63 & \cellcolor{bestbg}99.80 \\
C12 & 98.33 & \cellcolor{bestbg}99.25 & \cellcolor{bestbg}99.64 & \cellcolor{bestbg}99.77 \\
C16 & 99.99 & 99.86 & 99.92 & 99.95 \\
C18 & 98.01 & \cellcolor{bestbg}98.96 & \cellcolor{bestbg}99.49 & \cellcolor{bestbg}99.69 \\
C19 & 97.85 & \cellcolor{bestbg}99.06 & \cellcolor{bestbg}99.45 & \cellcolor{bestbg}99.62 \\
I3 & 93.67 & \cellcolor{bestbg}96.74 & \cellcolor{bestbg}97.91 & \cellcolor{bestbg}98.42 \\
I6 & 96.80 & \cellcolor{bestbg}98.69 & \cellcolor{bestbg}99.26 & \cellcolor{bestbg}99.52 \\
I8 & 96.82 & \cellcolor{bestbg}99.02 & \cellcolor{bestbg}99.45 & \cellcolor{bestbg}99.64 \\
I9 & 95.60 & \cellcolor{bestbg}98.14 & \cellcolor{bestbg}98.98 & \cellcolor{bestbg}99.21 \\
I10 & 96.04 & \cellcolor{bestbg}98.62 & \cellcolor{bestbg}99.38 & \cellcolor{bestbg}99.54 \\
\bottomrule
\end{tabular}
&
\begin{tabular}[t]{@{}l c ccc@{}}
\multicolumn{5}{c}{\textbf{$\ell_\infty$}} \\
\toprule
ID & AA & $\ens_{4\mathrm{k}}$ & $\ens_{8\mathrm{k}}$ & $\ens_{12\mathrm{k}}$ \\
\midrule
\multicolumn{5}{@{}l}{\itshape Train} \\
C1 & 99.63 & 99.59 & \cellcolor{bestbg}99.67 & \cellcolor{bestbg}99.70 \\
C2 & 99.72 & 99.56 & 99.71 & \cellcolor{bestbg}99.74 \\
C5 & 99.87 & 99.58 & 99.65 & 99.71 \\
C7 & 99.59 & 99.52 & \cellcolor{bestbg}99.60 & \cellcolor{bestbg}99.63 \\
C8 & 99.59 & 99.57 & \cellcolor{bestbg}99.63 & \cellcolor{bestbg}99.69 \\
C9 & 99.45 & 99.42 & \cellcolor{bestbg}99.53 & \cellcolor{bestbg}99.55 \\
C11 & 99.53 & \cellcolor{bestbg}99.78 & \cellcolor{bestbg}99.82 & \cellcolor{bestbg}99.84 \\
C13 & 99.84 & 99.59 & 99.69 & 99.78 \\
C14 & 98.64 & \cellcolor{bestbg}99.13 & \cellcolor{bestbg}99.67 & \cellcolor{bestbg}99.74 \\
C15 & 99.32 & \cellcolor{bestbg}99.65 & \cellcolor{bestbg}99.77 & \cellcolor{bestbg}99.79 \\
C17 & 99.59 & \cellcolor{bestbg}99.72 & \cellcolor{bestbg}99.80 & \cellcolor{bestbg}99.81 \\
C20 & 99.67 & 99.62 & \cellcolor{bestbg}99.76 & \cellcolor{bestbg}99.81 \\
I1 & 99.61 & 99.30 & 99.44 & \cellcolor{bestbg}99.62 \\
I2 & 99.24 & 99.19 & \cellcolor{bestbg}99.31 & \cellcolor{bestbg}99.56 \\
I4 & 98.69 & 98.34 & \cellcolor{bestbg}98.84 & \cellcolor{bestbg}99.18 \\
I5 & 98.49 & \cellcolor{bestbg}98.71 & \cellcolor{bestbg}98.79 & \cellcolor{bestbg}99.22 \\
I7 & 99.42 & 99.21 & 99.30 & \cellcolor{bestbg}99.57 \\
\midrule
\multicolumn{5}{@{}l}{\itshape Held-out} \\
C3 & 99.62 & \cellcolor{bestbg}99.66 & \cellcolor{bestbg}99.74 & \cellcolor{bestbg}99.75 \\
C4 & 99.46 & \cellcolor{bestbg}99.57 & \cellcolor{bestbg}99.68 & \cellcolor{bestbg}99.69 \\
C6 & 99.46 & 99.44 & \cellcolor{bestbg}99.57 & \cellcolor{bestbg}99.63 \\
C10 & 99.00 & \cellcolor{bestbg}99.60 & \cellcolor{bestbg}99.72 & \cellcolor{bestbg}99.78 \\
C12 & 99.50 & \cellcolor{bestbg}99.50 & \cellcolor{bestbg}99.60 & \cellcolor{bestbg}99.63 \\
C16 & 99.52 & \cellcolor{bestbg}99.77 & \cellcolor{bestbg}99.84 & \cellcolor{bestbg}99.87 \\
C18 & 99.65 & 99.44 & 99.58 & 99.62 \\
C19 & 99.45 & \cellcolor{bestbg}99.58 & \cellcolor{bestbg}99.68 & \cellcolor{bestbg}99.71 \\
I3 & 98.31 & \cellcolor{bestbg}98.53 & \cellcolor{bestbg}99.00 & \cellcolor{bestbg}99.12 \\
I6 & 99.55 & 98.88 & 99.06 & 99.37 \\
I8 & 99.56 & 99.15 & 99.30 & 99.50 \\
I9 & 98.29 & \cellcolor{bestbg}98.69 & \cellcolor{bestbg}99.11 & \cellcolor{bestbg}99.24 \\
I10 & 98.62 & \cellcolor{bestbg}98.97 & \cellcolor{bestbg}99.25 & \cellcolor{bestbg}99.46 \\
\bottomrule
\end{tabular}
\\
\end{tabular}}
\end{table*}

\begin{table*}[!t]
\centering
\footnotesize
\renewcommand{\arraystretch}{0.9}
\setlength{\tabcolsep}{4pt}
\caption{\textbf{Our greedy ensembles}: \attackoptimality{} (\aoi, \%) per model as the query budget $\ens$ grows ($4$k$\to$$12$k); train and held-out models. $\ell_0$ has no AutoAttack baseline.}
\label{tab:ours-vs-aa-l0}
\resizebox{\textwidth}{!}{%
\begin{tabular}{@{} l S[table-format=2.1] S[table-format=2.1] S[table-format=2.1] @{\hspace{9pt}} l S[table-format=2.1] S[table-format=2.1] S[table-format=2.1] @{\hspace{14pt}} l S[table-format=2.1] S[table-format=2.1] S[table-format=2.1] @{\hspace{9pt}} l S[table-format=2.1] S[table-format=2.1] S[table-format=2.1] @{}}
\toprule
\multicolumn{8}{@{}c@{\hspace{14pt}}}{\itshape Train} & \multicolumn{8}{c@{}}{\itshape Held-out} \\
\cmidrule(r{12pt}){1-8}\cmidrule{9-16}
ID & {$\ens_{4\mathrm{k}}$} & {$\ens_{8\mathrm{k}}$} & {$\ens_{12\mathrm{k}}$} & ID & {$\ens_{4\mathrm{k}}$} & {$\ens_{8\mathrm{k}}$} & {$\ens_{12\mathrm{k}}$} & ID & {$\ens_{4\mathrm{k}}$} & {$\ens_{8\mathrm{k}}$} & {$\ens_{12\mathrm{k}}$} & ID & {$\ens_{4\mathrm{k}}$} & {$\ens_{8\mathrm{k}}$} & {$\ens_{12\mathrm{k}}$} \\
\midrule
C1  & 96.6 & 98.5 & 99.2 & C15 & 91.1 & 97.0 & 98.2 & C3  & 96.6 & 98.4 & 99.2 & C19 & 96.8 & 98.4 & 99.2 \\
C2  & 98.5 & 99.5 & 99.8 & C17 & 97.1 & 98.6 & 99.4 & C4  & 95.8 & 97.7 & 98.7 & I3  & 98.0 & 98.8 & 99.3 \\
C5  & 98.3 & 99.4 & 99.9 & C20 & 97.2 & 98.5 & 99.3 & C6  & 96.3 & 98.2 & 99.0 & I6  & 99.1 & 99.5 & 99.8 \\
C7  & 96.6 & 98.3 & 99.2 & I1  & 99.3 & 99.6 & 99.8 & C10 & 97.0 & 98.7 & 99.5 & I8  & 99.4 & 99.6 & 99.8 \\
C8  & 96.8 & 98.6 & 99.4 & I2  & 98.6 & 99.3 & 99.7 & C12 & 96.5 & 98.2 & 99.0 & I9  & 99.1 & 99.5 & 99.8 \\
C9  & 97.7 & 99.1 & 99.8 & I4  & 98.3 & 99.1 & 99.6 & C16 & 92.5 & 96.8 & 98.1 & I10 & 99.0 & 99.5 & 99.8 \\
C11 & 98.6 & 99.4 & 99.7 & I5  & 99.1 & 99.5 & 99.7 & C18 & 96.7 & 98.3 & 99.1 & & & & \\
C13 & 97.2 & 98.8 & 99.6 & I7  & 98.7 & 99.2 & 99.7 & & & & & & & & \\
C14 & 97.2 & 98.3 & 99.0 & & & & & & & & & & & & \\
\bottomrule
\end{tabular}
}
\end{table*}
In Table~\ref{tab:ours-vs-aa-l0}, we report the \aoi{} on $\ell_0$, where AA provides no baseline. 
The \aoi{} on held-out models at $12$k ranges from $98.1$--$99.8$ across all models, showing that the ensembles reach the frontier reliably even on a norm with no established benchmark to compare against.

\begin{figure}[!t]
    \centering
    \begin{subfigure}[t]{0.48\linewidth}
        \centering
        \vspace{0.2em}
        \includegraphics[width=\linewidth, keepaspectratio]{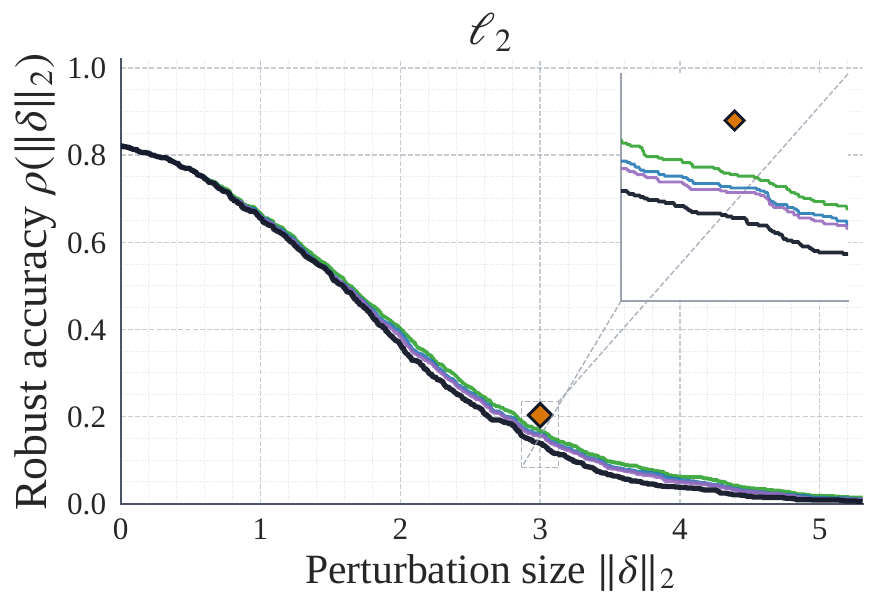}
        {\small (a) Best case: I3 on $\ell_2$}\\
            \end{subfigure}
    \hfill
    \begin{subfigure}[t]{0.48\linewidth}
        \centering
        \vspace{0.2em}
        \includegraphics[width=\linewidth, keepaspectratio]{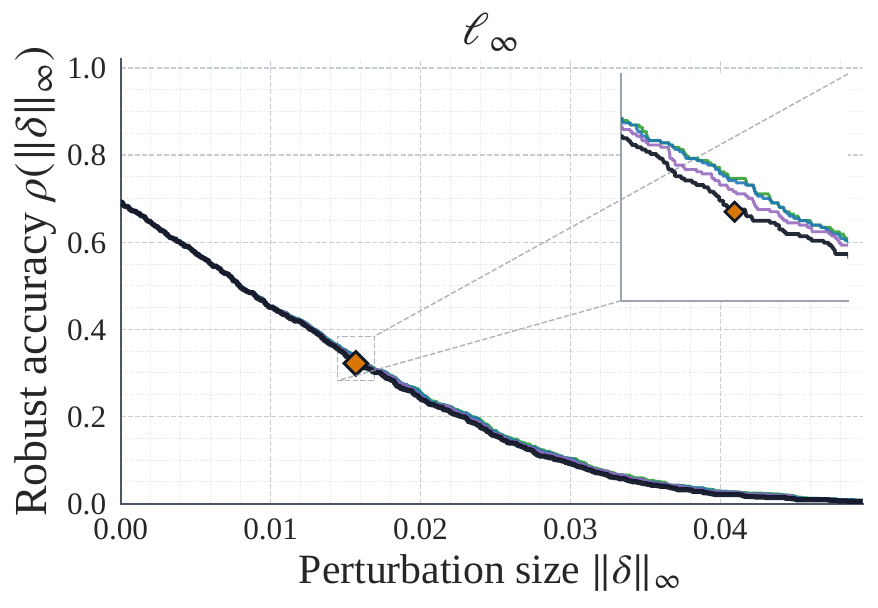}
        {\small (b) Worst case: I6 on $\ell_\infty$} \\
    \end{subfigure}

    \vspace{0.6em}

    \begin{subfigure}[t]{\linewidth}
        \centering
        \includegraphics[width=0.85\linewidth, keepaspectratio]{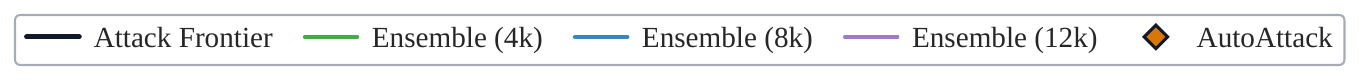}
    \end{subfigure}
    \caption{\textbf{Ensemble robustness curves vs.\ AA.} Our curve-based evaluations at $Q \in \{4\mathrm{k}, 8\mathrm{k}, 12\mathrm{k}\}$ compared against AA's point-wise baseline, on I3 (a) and I6 (b) as best and worst cases. Our ensemble wins in I3, while remaining competitive on I6.}
    \label{fig:frontier-cases}
\end{figure}

\begin{table*}[!t]
\centering
\scriptsize
\setlength{\tabcolsep}{4pt}
\caption{\textbf{Our greedy ensembles} vs.\ AutoAttack: Robust Accuracy $\rho(\varepsilon_{\text{ref}})$ (\%) at the reference $\varepsilon$ per model as the query budget grows ($4$k$\to$$12$k), for train and held-out models. $Q_{\mathrm{AA}}$ is the (variable) query cost of AA. \hlbest{Green}: our ensembles are at least as strong as AA.}
\label{tab:ours-vs-aa-ra}
\resizebox{\textwidth}{!}{%
\begin{tabular}{@{}c@{\hspace{10pt}}c@{\hspace{10pt}}c@{}}
\begin{tabular}[t]{@{}l S[table-format=2.1] S[table-format=4.0] S[table-format=2.1] S[table-format=2.1] S[table-format=2.1]@{}}
\multicolumn{6}{c}{\textbf{$\ell_1$}} \\
\toprule
ID & AA & $Q_{\mathrm{AA}}$ & $\ens_{4\mathrm{k}}$ & $\ens_{8\mathrm{k}}$ & $\ens_{12\mathrm{k}}$ \\
\midrule
\multicolumn{6}{@{}l}{\itshape Train} \\
C1 & 16.5 & 1824 & 17.2 & \cellcolor{bestbg}16.0 & \cellcolor{bestbg}15.4 \\
C2 & 31.0 & 3219 & \cellcolor{bestbg}30.8 & \cellcolor{bestbg}29.8 & \cellcolor{bestbg}29.2 \\
C5 & 26.2 & 2736 & 27.8 & 27.3 & 27.1 \\
C7 & 11.1 & 1286 & 12.4 & 12.0 & 11.7 \\
C8 & 52.0 & 5179 & \cellcolor{bestbg}52.0 & \cellcolor{bestbg}51.7 & \cellcolor{bestbg}51.7 \\
C9 & 41.3 & 4195 & \cellcolor{bestbg}40.7 & \cellcolor{bestbg}40.5 & \cellcolor{bestbg}40.3 \\
C11 & 43.1 & 4362 & 44.5 & 43.8 & 43.6 \\
C13 & 34.4 & 3535 & 36.5 & 34.9 & 34.5 \\
C14 & 0.0 & 198 & \cellcolor{bestbg}0.0 & \cellcolor{bestbg}0.0 & \cellcolor{bestbg}0.0 \\
C15 & 0.0 & 187 & \cellcolor{bestbg}0.0 & \cellcolor{bestbg}0.0 & \cellcolor{bestbg}0.0 \\
C17 & 8.5 & 1051 & 8.8 & \cellcolor{bestbg}7.6 & \cellcolor{bestbg}7.5 \\
C20 & 9.5 & 1174 & 10.9 & 9.9 & \cellcolor{bestbg}9.5 \\
I1 & 8.6 & 1015 & 9.1 & \cellcolor{bestbg}8.2 & \cellcolor{bestbg}8.1 \\
I2 & 36.0 & 3700 & 37.8 & 36.5 & 36.1 \\
I4 & 40.9 & 4160 & 42.5 & 41.1 & \cellcolor{bestbg}40.7 \\
I5 & 3.8 & 508 & 4.9 & 4.4 & 4.1 \\
I7 & 1.7 & 299 & 2.1 & 1.9 & 1.9 \\
\midrule
\multicolumn{6}{@{}l}{\itshape Held-out} \\
C3 & 10.7 & 1271 & 11.4 & \cellcolor{bestbg}9.7 & \cellcolor{bestbg}9.5 \\
C4 & 10.3 & 1230 & 10.8 & \cellcolor{bestbg}10.0 & \cellcolor{bestbg}9.5 \\
C6 & 9.1 & 1100 & \cellcolor{bestbg}8.7 & \cellcolor{bestbg}8.2 & \cellcolor{bestbg}7.8 \\
C10 & 5.3 & 733 & 7.0 & \cellcolor{bestbg}4.9 & \cellcolor{bestbg}4.4 \\
C12 & 14.7 & 1644 & 16.5 & 15.3 & 15.0 \\
C16 & 0.2 & 219 & 0.3 & 0.3 & 0.3 \\
C18 & 4.7 & 667 & 6.2 & 5.3 & \cellcolor{bestbg}4.7 \\
C19 & 11.3 & 1324 & \cellcolor{bestbg}11.1 & \cellcolor{bestbg}10.7 & \cellcolor{bestbg}10.4 \\
I3 & 34.5 & 3590 & 35.2 & \cellcolor{bestbg}32.9 & \cellcolor{bestbg}32.0 \\
I6 & 6.8 & 821 & 7.6 & 6.9 & \cellcolor{bestbg}6.7 \\
I8 & 3.0 & 453 & 4.0 & 3.5 & 3.2 \\
I9 & 36.1 & 3682 & 38.2 & 36.6 & \cellcolor{bestbg}35.7 \\
I10 & 35.2 & 3605 & 36.4 & \cellcolor{bestbg}35.0 & \cellcolor{bestbg}34.6 \\
\bottomrule
\end{tabular}
&
\begin{tabular}[t]{@{}l S[table-format=2.1] S[table-format=4.0] S[table-format=2.1] S[table-format=2.1] S[table-format=2.1]@{}}
\multicolumn{6}{c}{\textbf{$\ell_2$}} \\
\toprule
ID & AA & $Q_{\mathrm{AA}}$ & $\ens_{4\mathrm{k}}$ & $\ens_{8\mathrm{k}}$ & $\ens_{12\mathrm{k}}$ \\
\midrule
\multicolumn{6}{@{}l}{\itshape Train} \\
C1 & 58.4 & 5767 & \cellcolor{bestbg}58.3 & \cellcolor{bestbg}58.3 & \cellcolor{bestbg}58.3 \\
C2 & 73.4 & 7217 & 73.5 & 73.5 & \cellcolor{bestbg}73.4 \\
C5 & 70.3 & 6918 & 70.5 & 70.4 & 70.4 \\
C7 & 65.8 & 6486 & \cellcolor{bestbg}65.7 & \cellcolor{bestbg}65.6 & \cellcolor{bestbg}65.6 \\
C8 & 62.2 & 6116 & \cellcolor{bestbg}62.2 & \cellcolor{bestbg}62.2 & \cellcolor{bestbg}62.2 \\
C9 & 58.9 & 5807 & \cellcolor{bestbg}58.9 & \cellcolor{bestbg}58.9 & \cellcolor{bestbg}58.9 \\
C11 & 77.1 & 7566 & \cellcolor{bestbg}77.1 & \cellcolor{bestbg}77.1 & \cellcolor{bestbg}77.1 \\
C13 & 73.3 & 7205 & \cellcolor{bestbg}73.3 & \cellcolor{bestbg}73.3 & \cellcolor{bestbg}73.3 \\
C14 & 0.0 & 195 & \cellcolor{bestbg}0.0 & \cellcolor{bestbg}0.0 & \cellcolor{bestbg}0.0 \\
C15 & 0.9 & 275 & 1.0 & 1.0 & \cellcolor{bestbg}0.9 \\
C17 & 64.6 & 6377 & \cellcolor{bestbg}63.5 & \cellcolor{bestbg}63.5 & \cellcolor{bestbg}63.3 \\
C20 & 69.2 & 6828 & \cellcolor{bestbg}68.5 & \cellcolor{bestbg}68.4 & \cellcolor{bestbg}68.2 \\
I1 & 6.5 & 799 & \cellcolor{bestbg}5.6 & \cellcolor{bestbg}4.9 & \cellcolor{bestbg}4.8 \\
I2 & 27.5 & 2823 & \cellcolor{bestbg}25.9 & \cellcolor{bestbg}25.6 & \cellcolor{bestbg}25.6 \\
I4 & 34.3 & 3488 & \cellcolor{bestbg}32.3 & \cellcolor{bestbg}31.6 & \cellcolor{bestbg}31.3 \\
I5 & 2.6 & 384 & \cellcolor{bestbg}2.3 & \cellcolor{bestbg}2.1 & \cellcolor{bestbg}2.1 \\
I7 & 1.4 & 266 & \cellcolor{bestbg}1.0 & \cellcolor{bestbg}1.0 & \cellcolor{bestbg}1.0 \\
\midrule
\multicolumn{6}{@{}l}{\itshape Held-out} \\
C3 & 64.9 & 6402 & \cellcolor{bestbg}64.2 & \cellcolor{bestbg}64.2 & \cellcolor{bestbg}64.1 \\
C4 & 70.9 & 6983 & \cellcolor{bestbg}70.2 & \cellcolor{bestbg}70.1 & \cellcolor{bestbg}70.1 \\
C6 & 64.2 & 6338 & \cellcolor{bestbg}63.4 & \cellcolor{bestbg}63.0 & \cellcolor{bestbg}62.7 \\
C10 & 65.7 & 6473 & \cellcolor{bestbg}65.6 & \cellcolor{bestbg}65.5 & \cellcolor{bestbg}65.3 \\
C12 & 67.3 & 6633 & \cellcolor{bestbg}67.0 & \cellcolor{bestbg}66.7 & \cellcolor{bestbg}66.7 \\
C16 & 7.4 & 903 & \cellcolor{bestbg}7.4 & \cellcolor{bestbg}7.4 & \cellcolor{bestbg}7.4 \\
C18 & 59.3 & 5872 & \cellcolor{bestbg}58.7 & \cellcolor{bestbg}58.7 & \cellcolor{bestbg}58.6 \\
C19 & 71.1 & 7006 & \cellcolor{bestbg}70.5 & \cellcolor{bestbg}70.3 & \cellcolor{bestbg}70.3 \\
I3 & 20.3 & 2204 & \cellcolor{bestbg}16.8 & \cellcolor{bestbg}15.7 & \cellcolor{bestbg}15.5 \\
I6 & 4.3 & 578 & \cellcolor{bestbg}4.1 & \cellcolor{bestbg}3.4 & \cellcolor{bestbg}3.4 \\
I8 & 2.2 & 372 & \cellcolor{bestbg}2.0 & \cellcolor{bestbg}1.6 & \cellcolor{bestbg}1.6 \\
I9 & 31.6 & 3231 & \cellcolor{bestbg}28.9 & \cellcolor{bestbg}27.6 & \cellcolor{bestbg}27.4 \\
I10 & 31.4 & 3196 & \cellcolor{bestbg}28.9 & \cellcolor{bestbg}27.7 & \cellcolor{bestbg}27.5 \\
\bottomrule
\end{tabular}
&
\begin{tabular}[t]{@{}l S[table-format=2.1] S[table-format=4.0] S[table-format=2.1] S[table-format=2.1] S[table-format=2.1]@{}}
\multicolumn{6}{c}{\textbf{$\ell_\infty$}} \\
\toprule
ID & AA & $Q_{\mathrm{AA}}$ & $\ens_{4\mathrm{k}}$ & $\ens_{8\mathrm{k}}$ & $\ens_{12\mathrm{k}}$ \\
\midrule
\multicolumn{6}{@{}l}{\itshape Train} \\
C1 & 51.1 & 5073 & 51.2 & 51.2 & 51.2 \\
C2 & 37.6 & 3822 & 38.1 & 37.9 & 37.8 \\
C5 & 29.6 & 3031 & 30.2 & 30.1 & 30.1 \\
C7 & 58.6 & 5799 & \cellcolor{bestbg}58.6 & \cellcolor{bestbg}58.6 & \cellcolor{bestbg}58.5 \\
C8 & 35.5 & 3567 & 35.7 & 35.7 & 35.7 \\
C9 & 35.5 & 3587 & 36.0 & 36.0 & 36.0 \\
C11 & 44.1 & 4415 & \cellcolor{bestbg}44.0 & \cellcolor{bestbg}44.0 & \cellcolor{bestbg}44.0 \\
C13 & 39.2 & 3944 & 39.4 & 39.3 & \cellcolor{bestbg}39.2 \\
C14 & 0.0 & 195 & \cellcolor{bestbg}0.0 & \cellcolor{bestbg}0.0 & \cellcolor{bestbg}0.0 \\
C15 & 0.0 & 183 & \cellcolor{bestbg}0.0 & \cellcolor{bestbg}0.0 & \cellcolor{bestbg}0.0 \\
C17 & 62.5 & 6174 & 62.6 & 62.6 & 62.6 \\
C20 & 64.8 & 6411 & \cellcolor{bestbg}64.7 & \cellcolor{bestbg}64.7 & \cellcolor{bestbg}64.7 \\
I1 & 45.2 & 4496 & 45.5 & 45.5 & 45.3 \\
I2 & 46.0 & 4574 & 46.1 & 46.1 & 46.1 \\
I4 & 51.3 & 5077 & \cellcolor{bestbg}51.3 & \cellcolor{bestbg}51.3 & \cellcolor{bestbg}51.3 \\
I5 & 27.7 & 2794 & \cellcolor{bestbg}27.4 & \cellcolor{bestbg}27.4 & \cellcolor{bestbg}27.2 \\
I7 & 29.5 & 2961 & 30.1 & 30.0 & 29.7 \\
\midrule
\multicolumn{6}{@{}l}{\itshape Held-out} \\
C3 & 57.9 & 5733 & 58.0 & \cellcolor{bestbg}57.9 & \cellcolor{bestbg}57.9 \\
C4 & 67.8 & 6687 & 67.9 & 67.9 & 67.9 \\
C6 & 66.5 & 6558 & 66.7 & 66.7 & 66.7 \\
C10 & 57.6 & 5701 & \cellcolor{bestbg}57.5 & \cellcolor{bestbg}57.5 & \cellcolor{bestbg}57.4 \\
C12 & 61.9 & 6114 & \cellcolor{bestbg}61.9 & \cellcolor{bestbg}61.9 & \cellcolor{bestbg}61.9 \\
C16 & 0.0 & 192 & \cellcolor{bestbg}0.0 & \cellcolor{bestbg}0.0 & \cellcolor{bestbg}0.0 \\
C18 & 51.0 & 5074 & 51.6 & 51.5 & 51.5 \\
C19 & 67.8 & 6691 & \cellcolor{bestbg}67.8 & \cellcolor{bestbg}67.8 & \cellcolor{bestbg}67.8 \\
I3 & 53.9 & 5347 & 54.1 & 54.1 & 54.1 \\
I6 & 32.2 & 3244 & 33.7 & 33.6 & 33.1 \\
I8 & 39.1 & 3904 & 39.9 & 39.8 & 39.6 \\
I9 & 53.8 & 5320 & 54.0 & 54.0 & 54.0 \\
I10 & 53.0 & 5244 & 53.3 & 53.2 & 53.1 \\
\bottomrule
\end{tabular}
\\
\end{tabular}}
\end{table*}

\begin{figure}[!t]
  \centering
  \includegraphics[width=0.475\linewidth]{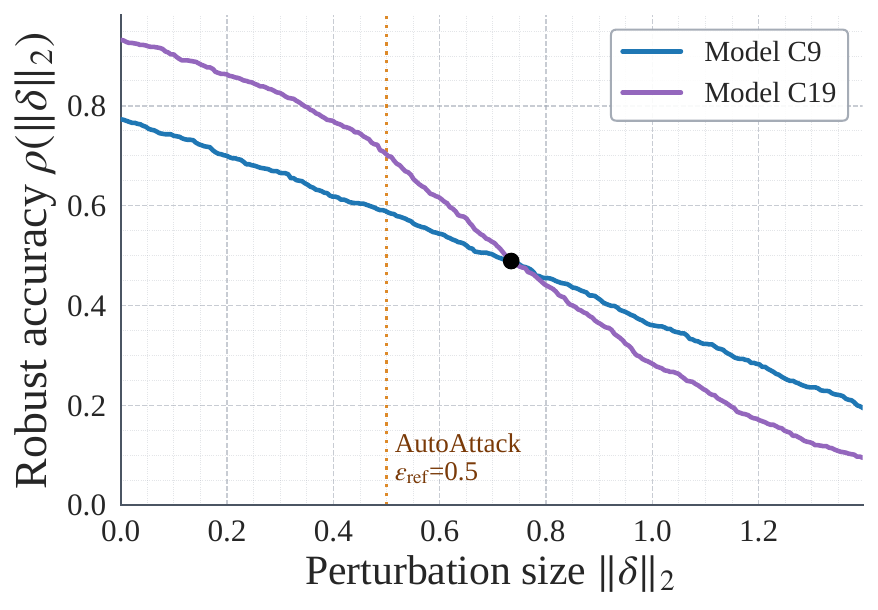}\hspace{0.03\linewidth}%
  \includegraphics[width=0.475\linewidth]{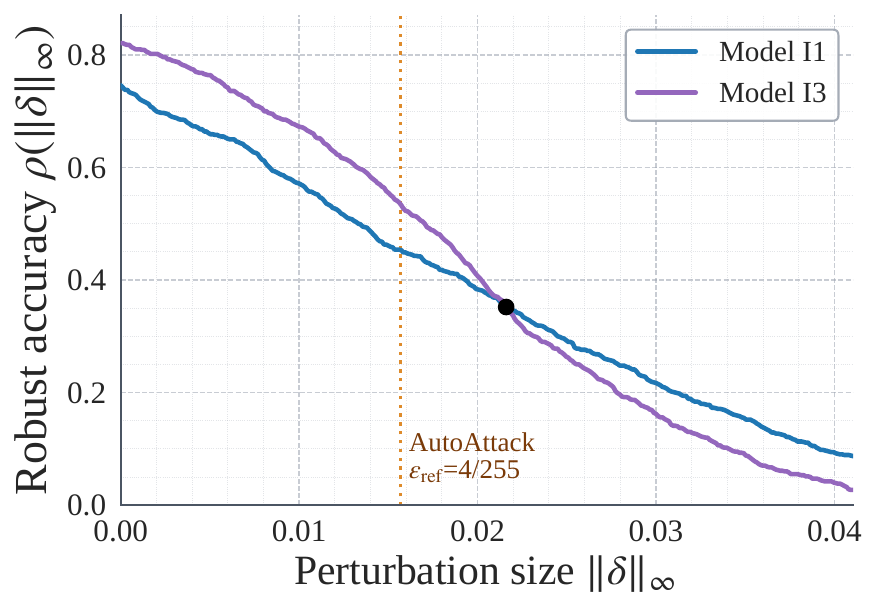}
  \caption{\textbf{Curves can cross past the reference $\varepsilon$.} Robust accuracy $\rho(\varepsilon)$ vs.\ perturbation budget for defense pairs whose curves intersect at a budget \emph{larger} than AA's reference $\varepsilon$ (dashed line): at the reference one defense leads, beyond it the order flips. Left: $\ell_2$ on CIFAR-10 (C9/C19); right: the single $\ell_\infty$ case we observe, on ImageNet (I1/I3).}
  \label{fig:crossings}
\end{figure}

We also examine how much increasing the query budget improves \aoi{} on the held-out pool of models, from $4$k to $12$k queries. 
On $\ell_1$, the mean \aoi{} rises from $97.83\pm0.53$ to $98.94\pm0.23$ ($+1.12$), the largest gain among the three norms.
On $\ell_\infty$, the gain is the smallest, from $99.29\pm0.39$ to $99.57\pm0.21$ ($+0.28$). This likely reflects $\ell_\infty$'s prevalence in robustness evaluation and adversarial training among the studied defenses: its attack configurations are near saturation even at low budgets.
$\ell_2$ falls in between, improving from $98.83\pm0.71$ to $99.56\pm0.37$ ($+0.73$), while retaining the largest gap to AA.

\myparagraph{Robust accuracy comparison.}
Here we directly compare our ensembles fixed-$\varepsilon$ performance at AA's reference value. This favors AA's fixed-$\varepsilon$ design because our allocation maximizes \gaoi{} over complete curves on the $17$ training models rather than optimizing specifically at $\varepsilon_{\mathrm{ref}}$, which can disadvantage our ensembles at low budgets. Nevertheless, our ensembles remain competitive at these reference values, as detailed in Table~\ref{tab:ours-vs-aa-ra} and illustrated by the best- and worst-case frontiers from the held-out models (Figure~\ref{fig:frontier-cases}).

$\ell_2$ is the norm where we perform best: our ensembles match or exceed AA on all $13$ held-out models, at every query tier, with a mean improvement of $1.5\pm1.6$ points at $12$k (positive values indicate a lower, hence better, robust accuracy than AA). 
$\ell_1$ shows instead a mixed picture: at $8$k queries we already match or exceed AA on $7$ out of $13$ held-out models, rising to $10$ out of $13$ at $12$k, with a mean improvement of $0.6\pm0.7$ points at $12$k.
This is in line with what we observed in the optimality analysis, where $\ell_1$ was the norm benefiting the most from additional queries. 
$\ell_\infty$ is instead the hardest case for us: we match or exceed AA on $5$ out of $13$ held-out models, at both $8$k and $12$k, with a mean difference of $-0.2\pm0.3$ points in AA's favor at $12$k.
The same pattern holds on the complete pool of $30$ models: at $12$k queries, we match or exceed AA on $29$ defenses on $\ell_2$, $20$ on $\ell_1$, and $13$ on $\ell_\infty$.

\begin{figure*}[!t]
    \centering
    \includegraphics[width=0.94\linewidth, keepaspectratio]{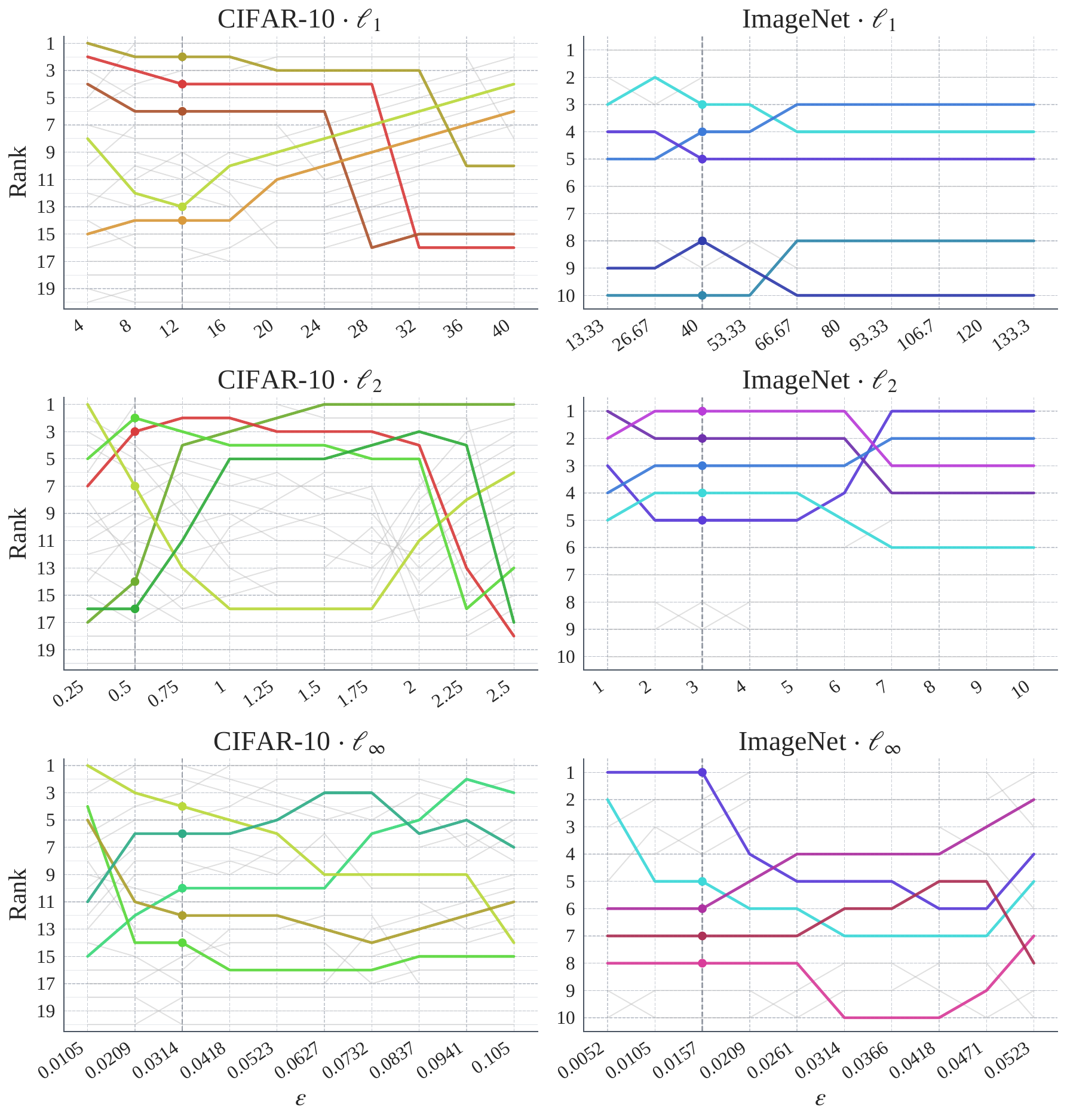}
    \vspace{0.3em}
    \includegraphics[width=0.62\linewidth, keepaspectratio]{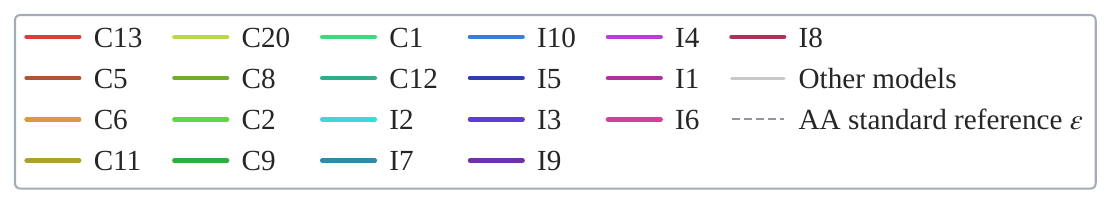}
    \caption{\textbf{Rank instability across perturbation budgets.} Defense rankings (y-axis) computed via AA across ten different perturbation sizes $\varepsilon$ (x-axis). In each panel, colored traces highlight the five models exhibiting the highest rank instability, while the remaining models are shown in gray. The vertical dashed line indicates the standard benchmark $\varepsilon$.
}
    \label{fig:rank-instability-all}
\end{figure*}

\begin{figure*}[!t]
    \centering
    \begin{subfigure}[t]{\linewidth}
        \begin{minipage}[c]{0.02\linewidth}
            \centering
            \rotatebox{90}{\small (a) CIFAR-10}
        \end{minipage}%
        \hfill
        \begin{minipage}[c]{0.97\linewidth}
            \centering
            \includegraphics[width=0.80\linewidth]{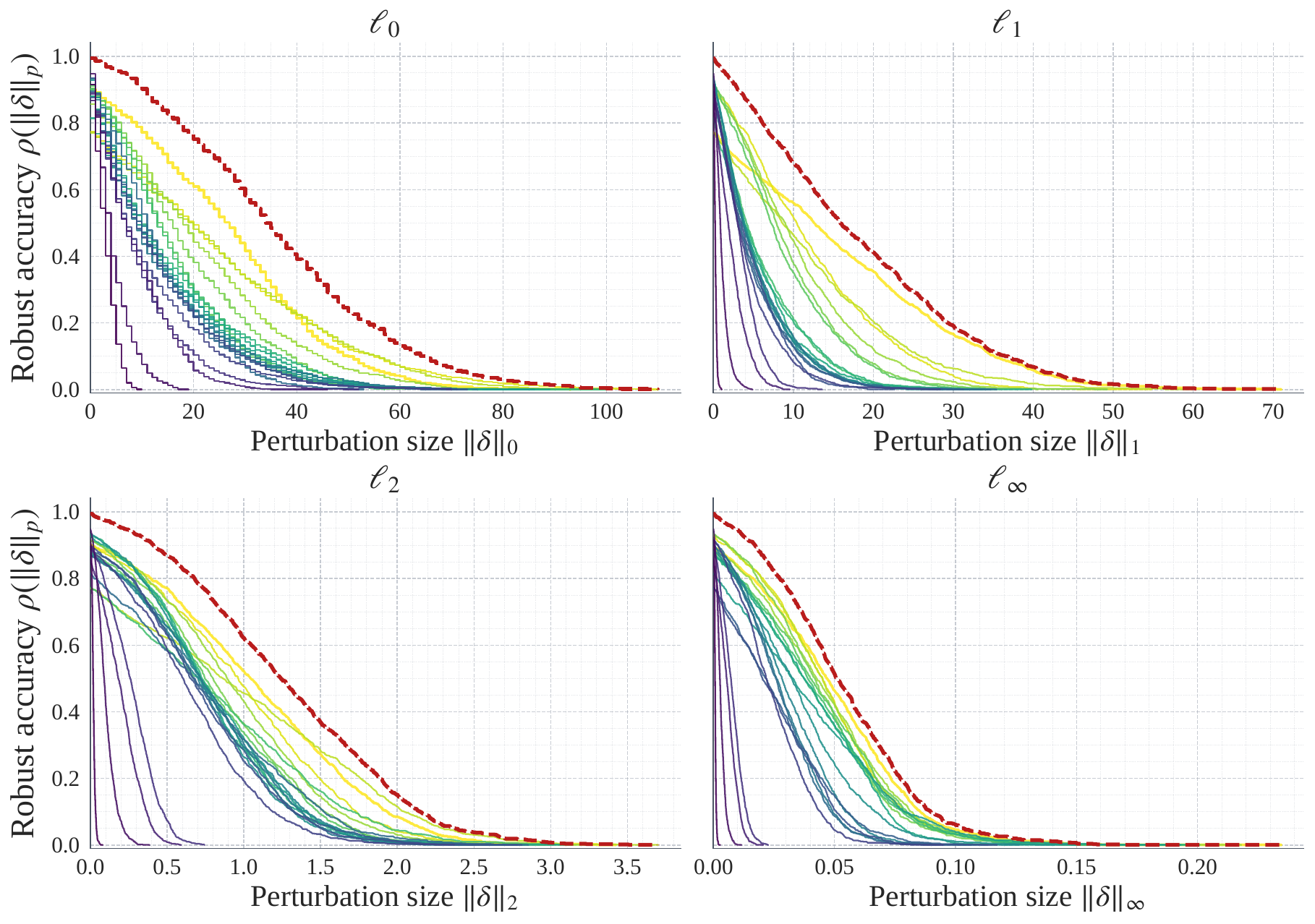}
        \end{minipage}
    \end{subfigure}
    \vspace{0.1em}
    \begin{subfigure}[t]{\linewidth}
        \begin{minipage}[c]{0.02\linewidth}
            \centering
            \rotatebox{90}{\small (b) ImageNet}
        \end{minipage}%
        \hfill
        \begin{minipage}[c]{0.97\linewidth}
            \centering
            \includegraphics[width=0.80\linewidth]{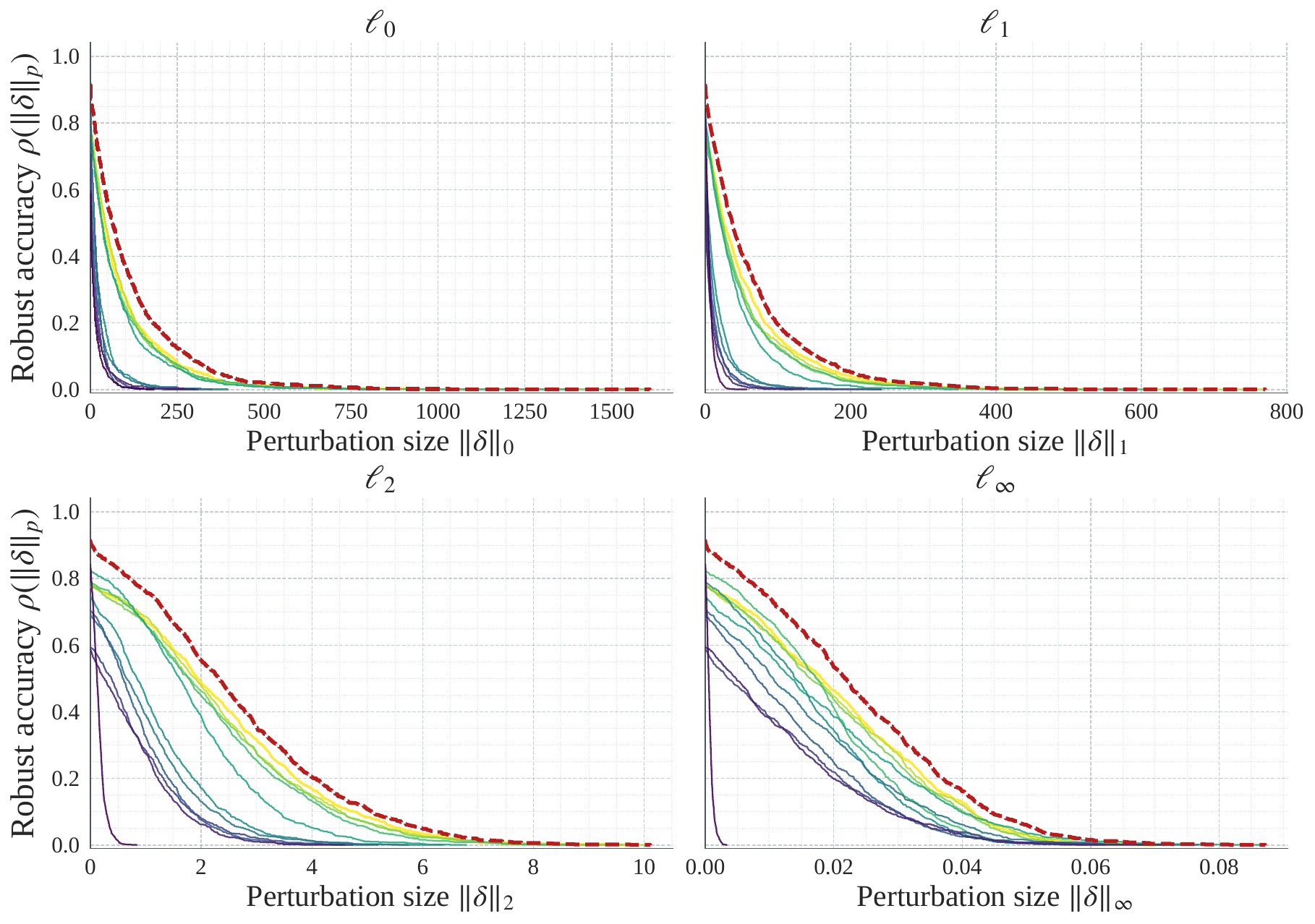}
        \end{minipage}
    \end{subfigure}
    \vspace{0.1em}
    \begin{subfigure}[t]{\linewidth}
        \centering
        \includegraphics[width=0.40\linewidth]{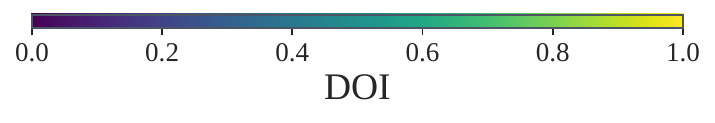}
    \end{subfigure}
    \caption{\textbf{Robustness--perturbation curves} for all $30$ evaluated defenses ($\ens_{12\mathrm{k}}$). For each dataset, the top row shows the $\ell_0$ and $\ell_1$ norms, while the bottom row shows $\ell_2$ and $\ell_\infty$. The curves plot the robust accuracy against the perturbation size, colored by their \doi{} score. The red dashed line represents the defense frontier.}
    \label{fig:defense-frontiers}
\end{figure*}

Beyond point-wise robust accuracy, the methods offer different control over cost and evaluation quality. AA's realized cost varies substantially across the held-out models, ranging from $219$ to $3\,682$ queries on $\ell_1$, from $372$ to $7\,006$ on $\ell_2$, and from $192$ to $6\,691$ on $\ell_\infty$, whereas our tiers use the predefined budgets $Q$. At AA's reference $\varepsilon$, the ensemble advantage is smaller than in the curve-based comparison: at $12$k queries, our ensembles match or exceed AA on $28$ of $39$ held-out model--norm pairs ($62$ of $90$ across the full pool).

\myparagraph{Crossings and ranking instability.}
Figure~\ref{fig:crossings} shows two observed cases in which a pair of defenses reverses order only beyond AA's reference $\varepsilon$: the ranking at the standard operating point therefore need not characterize their relative robustness over the complete curve. To assess the effect on rankings, we recompute AA's ordering over a grid of perturbation budgets (Figure~\ref{fig:rank-instability-all}). The changes are substantial: under $\ell_2$ on CIFAR-10, $17$ defenses move by at least five ranks across the evaluated range of perturbation budgets, while under $\ell_\infty$ on ImageNet, all $10$ models move by at least two ranks. Thus, current benchmarks that rank defenses by robust accuracy at one fixed reference $\varepsilon$, including AA, can miss ranking changes that arise across the complete robustness--perturbation curves.

\begin{table*}[!t]
\centering
\caption{\textbf{DOI Evaluations.} \doi{} computed with $\mathcal{E}$ at 4k, 8k, and 12k queries. The minimal variation between 4k and 12k demonstrates that the \doi{} provides stable robustness estimates.}
\label{tab:doi_stability}
\resizebox{\textwidth}{!}{
\begin{tabular}{@{} lSS>{\bfseries}S @{\hspace{1.5em}} lSS>{\bfseries}S @{\hspace{1.5em}} lSS>{\bfseries}S @{\hspace{1.5em}} lSS>{\bfseries}S @{}}
\toprule
\multicolumn{4}{c}{\textbf{$\ell_0$ Norm}} & \multicolumn{4}{c}{\textbf{$\ell_1$ Norm}} & \multicolumn{4}{c}{\textbf{$\ell_2$ Norm}} & \multicolumn{4}{c}{\textbf{$\ell_\infty$ Norm}} \\
\cmidrule(r){1-4} \cmidrule(lr){5-8} \cmidrule(lr){9-12} \cmidrule(l){13-16}
ID & {4k} & {8k} & \multicolumn{1}{c}{12k} & ID & {4k} & {8k} & \multicolumn{1}{c}{12k} & ID & {4k} & {8k} & \multicolumn{1}{c}{12k} & ID & {4k} & {8k} & \multicolumn{1}{c}{12k} \\
\midrule
\multicolumn{16}{c}{\textit{CIFAR-10}} \\
\midrule
C16 & 72.6 & 72.2 & 72.2 & C8 & 81.5 & 81.7 & 81.9 & C11 & 81.7 & 82.0 & 82.0 & C6 & 87.8 & 87.7 & 87.7 \\
C9 & 60.8 & 63.9 & 64.6 & C11 & 64.7 & 64.1 & 64.1 & C13 & 74.2 & 74.3 & 74.2 & C4 & 83.7 & 83.7 & 83.8 \\
C8 & 59.8 & 62.6 & 63.4 & C9 & 59.8 & 60.0 & 60.0 & C8 & 73.7 & 73.8 & 73.9 & C19 & 83.5 & 83.6 & 83.7 \\
C15 & 60.3 & 57.0 & 56.8 & C13 & 55.5 & 55.1 & 55.0 & C2 & 70.3 & 70.4 & 70.4 & C17 & 81.5 & 81.5 & 81.7 \\
C11 & 46.2 & 49.2 & 50.1 & C2 & 49.8 & 49.3 & 49.2 & C5 & 63.5 & 63.7 & 63.7 & C20 & 79.5 & 79.4 & 79.5 \\
C2 & 40.4 & 42.2 & 42.9 & C5 & 46.6 & 46.4 & 46.4 & C12 & 60.8 & 60.6 & 60.5 & C12 & 77.9 & 77.9 & 78.0 \\
C13 & 40.1 & 42.0 & 42.5 & C12 & 32.2 & 31.8 & 31.6 & C9 & 60.3 & 60.4 & 60.4 & C7 & 75.1 & 75.2 & 75.2 \\
C12 & 40.4 & 41.3 & 41.2 & C1 & 31.3 & 30.7 & 30.3 & C19 & 59.7 & 59.6 & 59.5 & C3 & 74.3 & 74.3 & 74.5 \\
C1 & 37.4 & 37.8 & 37.8 & C7 & 29.7 & 29.3 & 29.1 & C4 & 58.5 & 58.4 & 58.3 & C10 & 70.6 & 70.6 & 70.7 \\
C7 & 36.9 & 37.8 & 37.7 & C19 & 28.7 & 28.2 & 28.0 & C7 & 58.1 & 57.9 & 57.9 & C1 & 69.2 & 69.2 & 69.3 \\
C4 & 37.0 & 37.6 & 37.4 & C20 & 28.8 & 28.1 & 27.8 & C20 & 57.5 & 57.3 & 57.3 & C18 & 62.4 & 62.4 & 62.5 \\
C19 & 35.8 & 36.4 & 36.3 & C4 & 28.3 & 27.8 & 27.6 & C3 & 57.1 & 57.0 & 57.0 & C11 & 54.3 & 54.4 & 54.5 \\
C6 & 35.8 & 36.1 & 36.0 & C3 & 28.2 & 27.7 & 27.4 & C10 & 56.7 & 56.4 & 56.4 & C13 & 49.6 & 49.6 & 49.6 \\
C5 & 33.4 & 35.3 & 35.8 & C6 & 26.1 & 25.7 & 25.5 & C1 & 56.1 & 56.0 & 56.0 & C2 & 48.9 & 48.9 & 49.0 \\
C20 & 34.3 & 35.3 & 35.2 & C17 & 25.6 & 25.2 & 24.9 & C17 & 56.0 & 55.7 & 55.7 & C9 & 45.5 & 45.6 & 45.6 \\
C3 & 33.8 & 34.3 & 34.3 & C10 & 25.2 & 24.5 & 24.2 & C6 & 54.8 & 54.3 & 54.2 & C8 & 43.2 & 43.3 & 43.3 \\
C17 & 29.5 & 30.2 & 30.1 & C18 & 23.1 & 22.5 & 22.2 & C18 & 48.9 & 48.5 & 48.5 & C5 & 43.0 & 43.1 & 43.2 \\
C18 & 24.2 & 24.7 & 24.7 & C16 & 17.2 & 17.3 & 17.3 & C16 & 19.8 & 19.9 & 19.9 & C16 & 13.4 & 13.5 & 13.5 \\
C10 & 22.7 & 23.4 & 23.5 & C15 & 11.7 & 11.7 & 11.7 & C15 & 14.4 & 14.4 & 14.4 & C15 & 10.2 & 10.3 & 10.3 \\
C14 & 7.8 & 8.3 & 8.4 & C14 & 1.3 & 1.3 & 1.2 & C14 & 1.5 & 1.5 & 1.5 & C14 & 1.3 & 1.3 & 1.3 \\
\addlinespace
\midrule
\multicolumn{16}{c}{\textit{ImageNet}} \\
\midrule
I4 & 73.3 & 73.7 & 73.8 & I4 & 82.3 & 82.0 & 81.6 & I4 & 86.3 & 86.3 & 86.3 & I9 & 83.8 & 83.7 & 83.9 \\
I3 & 70.7 & 71.6 & 72.6 & I9 & 74.7 & 74.5 & 74.1 & I9 & 82.7 & 82.9 & 82.8 & I10 & 82.0 & 82.0 & 82.0 \\
I9 & 66.0 & 67.5 & 68.5 & I2 & 70.1 & 71.0 & 71.5 & I10 & 80.4 & 80.6 & 80.6 & I4 & 79.6 & 79.5 & 79.5 \\
I10 & 65.9 & 66.8 & 68.2 & I10 & 69.4 & 69.2 & 69.3 & I2 & 77.2 & 78.1 & 78.2 & I3 & 75.1 & 74.9 & 75.0 \\
I2 & 61.0 & 62.1 & 63.5 & I3 & 58.3 & 57.8 & 57.6 & I3 & 66.9 & 66.9 & 66.7 & I1 & 73.8 & 74.0 & 74.1 \\
I1 & 19.9 & 20.7 & 21.3 & I1 & 21.9 & 21.8 & 21.9 & I1 & 40.9 & 41.2 & 41.2 & I2 & 65.6 & 65.8 & 65.9 \\
I6 & 17.1 & 17.5 & 18.0 & I6 & 18.2 & 18.1 & 18.2 & I6 & 35.8 & 35.9 & 35.9 & I8 & 61.8 & 61.9 & 61.9 \\
I8 & 12.0 & 12.5 & 12.8 & I8 & 13.1 & 13.0 & 12.9 & I8 & 30.3 & 30.5 & 30.5 & I6 & 51.0 & 51.1 & 50.9 \\
I5 & 11.3 & 11.6 & 11.9 & I5 & 12.4 & 12.3 & 12.3 & I5 & 26.5 & 26.7 & 26.7 & I7 & 44.6 & 44.7 & 44.6 \\
I7 & 8.5 & 8.8 & 8.9 & I7 & 9.5 & 9.6 & 9.6 & I7 & 23.9 & 24.1 & 24.1 & I5 & 44.0 & 44.1 & 43.7 \\
\bottomrule
\end{tabular}
}
\end{table*}

\myparagraph{\doi{} Rankings.}
Table~\ref{tab:doi_stability} reports the norm-specific \doi{} rankings across all query tiers. Because each ranking summarizes complete robustness--perturbation curves, it does not require selecting a reference $\varepsilon$ for that norm.

Comparing the resulting ranks reveals strong cross-norm asymmetry. C6~\cite{gowal2021improving}, adversarially trained on $\ell_\infty$, ranks first there but falls in the bottom half under the other norms. C15 and C16~\cite{zhong2024towards}, trained on $\ell_0$, show the converse pattern, while C8~\cite{jiang2023towards}, trained on $\ell_1$, ranks near the top under $\ell_0$, $\ell_1$, and $\ell_2$ but near the bottom under $\ell_\infty$. Thus, a high rank under one norm does not imply a high rank under the others.

Broad cross-norm robustness also does not follow directly from the norm used for adversarial training. Of the two ResNet-18 models C10 and
C11~\cite{rade2021helper}, C11 ranks fifth under $\ell_0$ and in the top two under $\ell_1$ and $\ell_2$, whereas C10 ranks $19$th under $\ell_0$ but ninth under $\ell_\infty$. C9~\cite{maini2020adversarial}, trained on the union of $\ell_1$, $\ell_2$, and $\ell_\infty$, ranks near the top under $\ell_0$ and $\ell_1$ but only $15$th under $\ell_\infty$. On ImageNet, I4, I9, and I10~\cite{singh2023revisiting} rank near the top under all four norms, and I2, I3, I4, I9, and I10 consistently form the top half under $\ell_0$, $\ell_1$, and $\ell_2$. These patterns are also visible in Figure~\ref{fig:defense-frontiers}.

The rankings are stable across the evaluated tiers: six of the eight (norm, dataset) combinations retain the same strict ordering at $4$k, $8$k, and $12$k. On CIFAR-10, $\ell_1$ differs only by a C19/C20 tie at $4$k, while $\ell_0$ contains small local swaps at $4$k that resolve by $8$k. This is empirical stability rather than a guarantee, since the frontier and \doi{} can change with the query budget, as seen under $\ell_0$.

\subsection{Using the ensembles in practice}

\begin{figure}[!b]
  \centering
  \includegraphics[width=0.9\linewidth]{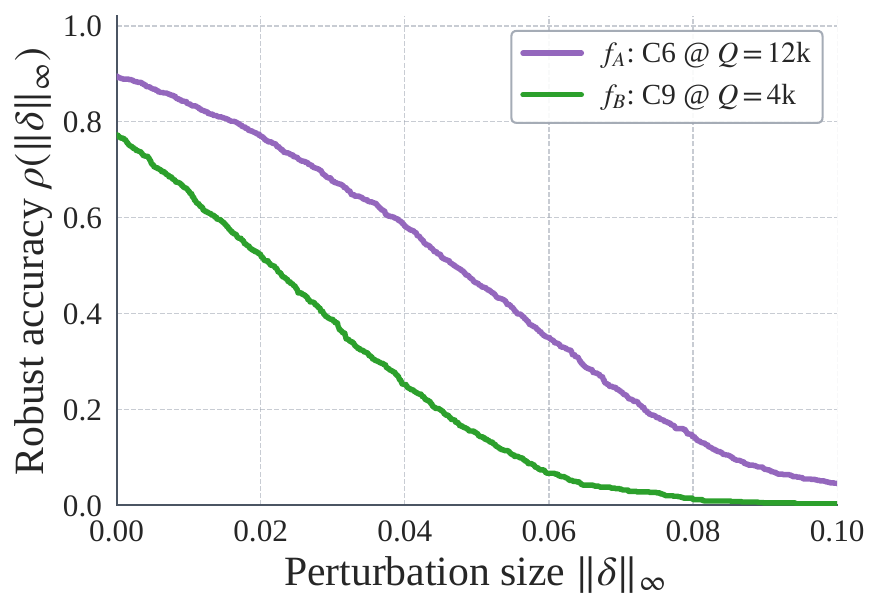}
  \caption{\textbf{Incremental evaluation with a controllable budget.} 
  Robustness--perturbation curves on $\ell_\infty$ for a defense 
  $f_A$ (C6) evaluated with the full $\ens_{12\mathrm{k}}$ ensemble and a 
  candidate $f_B$ (C9) evaluated with the cheaper $\ens_{4\mathrm{k}}$ tier. 
  The new curve is below the reference at 
  $4\mathrm{k}$ queries; more queries can only lower it further, 
  thus $f_B$ can be directly rejected.}
  \label{fig:practice-example}
\end{figure}

Suppose a practitioner has an established defense $f_A$, already evaluated with the full ensemble $\ens_{12\mathrm{k}}$, and wants to decide whether a new candidate $f_B$ improves on it. 
Rather than committing the full budget upfront, they can first attack $f_B$ with the cheapest tier $\ens_{4\mathrm{k}}$, on the same samples and norm, and compare the two robustness--perturbation 
curves. 
This early comparison is already conclusive in one direction. Since the ensemble distances upper-bound the true minimal perturbations, and adding queries can only tighten them, the curve of $f_B$ can only move downward as the budget grows from $4\mathrm{k}$ to $12\mathrm{k}$: spending more queries can only reveal $f_B$ to be weaker than it already appears, never stronger. 
Therefore, if the $\ens_{4\mathrm{k}}$ curve of $f_B$ already lies below the $\ens_{12\mathrm{k}}$ curve of $f_A$, no additional query can reverse the outcome, and $f_B$ can be rejected after only $4\mathrm{k}$ queries, without requiring further computations. 

Figure~\ref{fig:practice-example} shows one such case on $\ell_\infty$ taken from our model pool: the curve of C9 evaluated at $\ens_{4\mathrm{k}}$ already sits entirely below that of C6 evaluated at $\ens_{12\mathrm{k}}$, so C9 is less robust than C6 across every perturbation size, and the remaining $8\mathrm{k}$ queries would only widen the gap. 
The practitioner discards C9 having spent a third of the full budget. 
If instead the two curves are close or overlap at $4\mathrm{k}$, the comparison is not yet decisive: the practitioner can then resume the same run and extend it to $8\mathrm{k}$ or $12\mathrm{k}$, refining the estimate only where it is needed.

A complementary view for comparing the two models directly is to compare their $\mathrm{AUREC}$ (Eq.~\eqref{eq:aurec}). 
If the $\mathrm{AUREC}$ of $f_B$ at $4\mathrm{k}$ is already much smaller than that of $f_A$ at $12\mathrm{k}$, $f_B$ can be discarded on the same early-stopping grounds as above, since a smaller area already signals lower overall robustness before the estimate has even tightened. 
This view is complementary, not a substitute, for comparing the models directly: $\mathrm{AUREC}$ summarizes robustness as an average over the whole perturbation range, but it does not reveal whether a defense is 
more robust at small perturbations than at large ones, or vice versa. 
A practitioner should instead choose between the two defenses also by considering where along the curve robustness matters most for their specific use case.

\section{Related Work}\label{sec:related}

\myparagraph{Fixed-$\varepsilon$ Ensemble evaluation.}
Carlini et al.~\cite{carlini2019evaluating} showed that fixed-$\varepsilon$ evaluation is fragile: it can conflate attack failure with genuine robustness. They advocated for testing across multiple perturbation budgets and adapting attacks to defense-specific properties. 
AA~\cite{croce2020reliable} introduced a fixed, parameter-free ensemble of four attacks (Auto-Projected Gradient Descent with Cross-Entropy and Difference-of-Logits-Ratio losses, Fast Adaptive Boundary, and Square attacks), establishing the de facto standard for robustness benchmarking. Adaptive-AA~\cite{yao2021automated} extends AA by automatically tuning attack hyperparameters. Sparse-AA~\cite{zhong2024towards} applies the fixed-ensemble approach to pixel-wise $\ell_0$ sparsity, constrained to grouped-channel perturbations. Finally, AutoAE~\cite{liu2023autoae} proposes a greedy algorithm to allocate iterations across attacks at a single, fixed $\varepsilon$, thus it does not extend to curve-based evaluation.
We provide evidence of the brittleness of fixed-$\varepsilon$ evaluations in our experiments. Moreover, none of these methods covers the optimality of the algorithms used for the evaluations.

\myparagraph{Adversarial Benchmarks.}
RobustBench~\cite{croce2021robustbench} maintains standardized leaderboards reporting robust accuracy at fixed budgets on $\ell_2$ and $\ell_\infty$ norms via AA. Guo et al.~\cite{guo2023comprehensive} propose a multi-view evaluation framework with $23$ data- and model-oriented metrics, including test-example integrity, model structure, and adversarial behavior. AttackBench~\cite{cina2025attackbench} takes a complementary view, proposing the \attackoptimality{} (\aoi{}) to measure how closely individual attacks approach the smallest achievable perturbation across models. However, none of these two approaches propose a method for combining attacks.

\section{Conclusion and Future Work}\label{sec:conclusion}
We introduce a new framework for adversarial robustness evaluation. 
Our greedy algorithm selects a minimum-norm attack ensemble, including \APGDmin{}, within a query budget to estimate the complete robustness--perturbation curve. 
Its \aoi{} quantifies how closely the ensemble approaches the empirical attack frontier.
The ensembles generalize beyond the models used to construct them. Our $12$k-query ensembles match or exceed AA's \aoi{} in most cases. 
The \doi{} summarizes the complete curve relative to the empirical defense frontier, avoiding the selection of one $\varepsilon$ for the ranking. 
The nested tiers make both the query cost and the tightness of the frontier approximation controllable.

Our framework has three main limitations, each pointing to a natural extension. 
First, evaluations are pool-based: the frontiers are empirical estimates and the \doi{} is relative to the shared defense pool. We mitigate this by fixing and releasing the reference pool, so that new defenses are ranked against a common frontier. 
Second, constructing the frontier requires pre-computing distance--query trajectories for every attack on the reference models, an offline cost that $Q$ does not control. Cached trajectories make extensions incremental, since adding an attack only requires its own runs, and warm-starting new runs from existing ones would reduce this cost further.
Finally, the pool contains only white-box, gradient-based attacks, so defenses relying on gradient obfuscation or non-differentiable components are not faithfully evaluated. Since the min-composition framework only consumes per-sample distances, black-box attacks and semantic or patch-based perturbations can be added to the pool at the sole cost of generalizing the query unit.

We believe this work positions curve-based, optimality-aware, and configurable evaluation as a practical replacement for fixed-$\varepsilon$ leaderboards.

\section*{Acknowledgements}
This work has been carried out while L. Scionis and L. Melis were enrolled in the Italian National Doctorate on AI run by the Sapienza University of Rome in collaboration with the University of Cagliari.
This research has been partially supported by the Horizon Europe projects Sec4AI4Sec (GA no. 101120393), and CoEvolution (GA no. 101168560), and by project FISA-2023-00128 funded by the MUR program \textit{Fondo italiano per le scienze applicate}.

\myparagraph{Declaration on the use of generative AI and AI-assisted technologies.}
During the preparation of this work, the author(s) used Claude to complement the initial literature review and to write, and Claude Code to polish the plots and visualizations. The author(s) reviewed and edited the output as needed and take full responsibility for the content of the published article.

\bibliographystyle{elsarticle-num}

\begin{thebibliography}{10}
\expandafter\ifx\csname url\endcsname\relax
  \def\url#1{\texttt{#1}}\fi
\expandafter\ifx\csname urlprefix\endcsname\relax\def\urlprefix{URL }\fi
\expandafter\ifx\csname href\endcsname\relax
  \def\href#1#2{#2} \def\path#1{#1}\fi

\bibitem{biggio13-ecml}
B.~Biggio, I.~Corona, D.~Maiorca, B.~Nelson, N.~{\v{S}}rndi{\'c}, P.~Laskov, G.~Giacinto, F.~Roli, Evasion attacks against machine learning at test time, in: Machine Learning and Knowledge Discovery in Databases, Vol. 8190 of Lecture Notes in Computer Science, Springer, 2013, pp. 387--402.

\bibitem{szegedy2014intriguing}
C.~Szegedy, W.~Zaremba, I.~Sutskever, J.~Bruna, D.~Erhan, I.~J. Goodfellow, R.~Fergus, Intriguing properties of neural networks, in: International Conference on Learning Representations, 2014.

\bibitem{croce2021robustbench}
F.~Croce, M.~Andriushchenko, V.~Sehwag, E.~Debenedetti, N.~Flammarion, M.~Chiang, P.~Mittal, M.~Hein, {RobustBench}: A standardized adversarial robustness benchmark, in: NeurIPS Datasets and Benchmarks, 2021.

\bibitem{croce2020reliable}
F.~Croce, M.~Hein, Reliable evaluation of adversarial robustness with an ensemble of diverse parameter-free attacks, in: International Conference on Machine Learning, 2020.

\bibitem{biggio2018wild}
B.~Biggio, F.~Roli, Wild patterns: Ten years after the rise of adversarial machine learning, Pattern Recognition 84 (2018) 317--331.
\newblock \href {https://doi.org/https://doi.org/10.1016/j.patcog.2018.07.023} {\path{doi:https://doi.org/10.1016/j.patcog.2018.07.023}}.

\bibitem{carlini2019evaluating}
N.~Carlini, A.~Athalye, N.~Papernot, W.~Brendel, J.~Rauber, D.~Tsipras, I.~Goodfellow, A.~Madry, A.~Kurakin, On evaluating adversarial robustness, arXiv preprint arXiv:1902.06705 (2019).

\bibitem{risse2021compare}
N.~Risse, C.~G{\"o}pfert, J.~P. G{\"o}pfert, How to compare adversarial robustness of classifiers from a global perspective, in: International Conference on Artificial Neural Networks, Springer, 2021, pp. 29--41.

\bibitem{cina2025attackbench}
A.~E. Cin{\`a}, J.~Rony, M.~Pintor, L.~Demetrio, A.~Demontis, B.~Biggio, I.~{Ben Ayed}, F.~Roli, {AttackBench}: Evaluating gradient-based attacks for adversarial examples, Proceedings of the AAAI Conference on Artificial Intelligence 39~(3) (2025) 2600--2608.

\bibitem{gopfert2019adversarial}
C.~G{\"o}pfert, J.~P. G{\"o}pfert, B.~Hammer, Adversarial robustness curves, in: Joint European Conference on Machine Learning and Knowledge Discovery in Databases, Springer, 2019, pp. 172--179.

\bibitem{croce2022sparse}
F.~Croce, M.~Hein, Mind the box: {$\ell_1$}-{APGD} for sparse adversarial attacks on image classifiers, in: International Conference on Machine Learning, 2021.

\bibitem{pintor2021fast}
M.~Pintor, F.~Roli, W.~Brendel, B.~Biggio, Fast minimum-norm adversarial attacks through adaptive norm constraints, Advances in Neural Information Processing Systems 34 (2021) 20052--20062.

\bibitem{addepalli2022scaling}
S.~Addepalli, S.~Jain, G.~Sriramanan, R.~Venkatesh~Babu, Scaling adversarial training to large perturbation bounds, in: European Conference on Computer Vision, Springer, 2022, pp. 301--316.

\bibitem{rade2021helper}
R.~Rade, S.-M. Moosavi-Dezfooli, Helper-based adversarial training: Reducing excessive margin to achieve a better accuracy vs. robustness trade-off, in: ICML 2021 Workshop on Adversarial Machine Learning, 2021.

\bibitem{chen2024data}
E.-C. Chen, C.-R. Lee, Data filtering for efficient adversarial training, Pattern Recognition 151 (2024) 110394.

\bibitem{augustin2020adversarial}
M.~Augustin, A.~Meinke, M.~Hein, Adversarial robustness on in-and out-distribution improves explainability, in: European Conference on Computer Vision, Springer, 2020, pp. 228--245.

\bibitem{rebuffi2021fixing}
S.-A. Rebuffi, S.~Gowal, D.~A. Calian, F.~Stimberg, O.~Wiles, T.~Mann, Fixing data augmentation to improve adversarial robustness, arXiv preprint arXiv:2103.01946 (2021).

\bibitem{debenedetti2023light}
E.~Debenedetti, V.~Sehwag, P.~Mittal, A light recipe to train robust vision transformers, in: 2023 IEEE conference on secure and trustworthy machine learning (SaTML), IEEE, 2023, pp. 225--253.

\bibitem{sehwag2021robust}
V.~Sehwag, S.~Mahloujifar, T.~Handina, S.~Dai, C.~Xiang, M.~Chiang, P.~Mittal, Robust learning meets generative models: Can proxy distributions improve adversarial robustness?, arXiv preprint arXiv:2104.09425 (2021).

\bibitem{rodriguez2024characterizing}
A.~Rodr{\'\i}guez-Mu{\~n}oz, T.~Wang, A.~Torralba, Characterizing model robustness via natural input gradients, in: European Conference on Computer Vision, Springer, 2024, pp. 161--178.

\bibitem{cui2024decoupled}
J.~Cui, Z.~Tian, Z.~Zhong, X.~Qi, B.~Yu, H.~Zhang, Decoupled kullback-leibler divergence loss, Advances in Neural Information Processing Systems 37 (2024) 74461--74486.

\bibitem{stutz2020confidence}
D.~Stutz, M.~Hein, B.~Schiele, Confidence-calibrated adversarial training: Generalizing to unseen attacks, in: International conference on machine learning, PMLR, 2020, pp. 9155--9166.

\bibitem{singh2023revisiting}
N.~D. Singh, F.~Croce, M.~Hein, Revisiting adversarial training for imagenet: Architectures, training and generalization across threat models, Advances in Neural Information Processing Systems 36 (2023) 13931--13955.

\bibitem{zhong2024towards}
X.~Zhong, Y.~Huang, C.~Liu, Towards efficient training and evaluation of robust models against $\ell_0$ bounded adversarial perturbations, in: Forty-first International Conference on Machine Learning, 2024.

\bibitem{wong2020fast}
E.~Wong, L.~Rice, J.~Z. Kolter, Fast is better than free: Revisiting adversarial training, arXiv preprint arXiv:2001.03994 (2020).

\bibitem{gowal2021improving}
S.~Gowal, S.-A. Rebuffi, O.~Wiles, F.~Stimberg, D.~A. Calian, T.~A. Mann, Improving robustness using generated data, Advances in neural information processing systems 34 (2021) 4218--4233.

\bibitem{pang2022robustness}
T.~Pang, M.~Lin, X.~Yang, J.~Zhu, S.~Yan, Robustness and accuracy could be reconcilable by (proper) definition, in: International conference on machine learning, PMLR, 2022, pp. 17258--17277.

\bibitem{salman2020adversarially}
H.~Salman, A.~Ilyas, L.~Engstrom, A.~Kapoor, A.~Madry, Do adversarially robust imagenet models transfer better?, Advances in Neural Information Processing Systems 33 (2020) 3533--3545.

\bibitem{jiang2023towards}
Y.~Jiang, C.~Liu, Z.~Huang, M.~Salzmann, S.~Susstrunk, Towards stable and efficient adversarial training against $\ell_1$ bounded adversarial attacks, in: International Conference on Machine Learning, PMLR, 2023, pp. 15089--15104.

\bibitem{maini2020adversarial}
P.~Maini, E.~Wong, Z.~Kolter, Adversarial robustness against the union of multiple perturbation models, in: International Conference on Machine Learning, PMLR, 2020, pp. 6640--6650.

\bibitem{wang2023better}
Z.~Wang, T.~Pang, C.~Du, M.~Lin, W.~Liu, S.~Yan, Better diffusion models further improve adversarial training, in: International conference on machine learning, PMLR, 2023, pp. 36246--36263.

\bibitem{xu2023exploring}
Y.~Xu, Y.~Sun, M.~Goldblum, T.~Goldstein, F.~Huang, Exploring and exploiting decision boundary dynamics for adversarial robustness, arXiv preprint arXiv:2302.03015 (2023).

\bibitem{matyasko2021pdpgd}
A.~Matyasko, L.-P. Chau, {PDPGD}: Primal-dual proximal gradient descent adversarial attack, arXiv preprint arXiv:2106.01538 (2021).

\bibitem{rony2020augmented}
J.~Rony, E.~Granger, M.~Pedersoli, I.~{Ben Ayed}, Augmented lagrangian adversarial attacks, in: Proceedings of the IEEE/CVF International Conference on Computer Vision, 2021, pp. 7738--7747.

\bibitem{rony2019decoupling}
J.~Rony, L.~G. Hafemann, L.~S. Oliveira, I.~{Ben Ayed}, R.~Sabourin, E.~Granger, Decoupling direction and norm for efficient gradient-based {$\ell_2$} adversarial attacks and defenses, in: Proceedings of the IEEE/CVF Conference on Computer Vision and Pattern Recognition, 2019, pp. 4322--4330.

\bibitem{cina2025sigmazero}
A.~E. Cin{\`a}, F.~Villani, M.~Pintor, L.~Sch{\"o}nherr, B.~Biggio, M.~Pelillo, $\sigma$-zero: Gradient-based optimization of $\ell_0$-norm adversarial examples, in: The Thirteenth International Conference on Learning Representations, 2025.

\bibitem{yao2021automated}
C.~Yao, P.~Bielik, P.~Tsankov, M.~Vechev, Automated discovery of adaptive attacks on adversarial defenses, Advances in Neural Information Processing Systems 34 (2021) 26858--26870.

\bibitem{liu2023autoae}
S.~Liu, F.~Peng, K.~Tang, Reliable robustness evaluation via automatically constructed attack ensembles, in: Proceedings of the AAAI Conference on Artificial Intelligence (AAAI), 2023.

\bibitem{guo2023comprehensive}
J.~Guo, W.~Bao, J.~Wang, Y.~Ma, X.~Gao, G.~Xiao, A.~Liu, J.~Dong, X.~Liu, W.~Wu, A comprehensive evaluation framework for deep model robustness, Pattern Recognition 137 (2023) 109308.

\end{thebibliography}

\end{document}